\documentclass{article}
\usepackage[dvipsnames]{xcolor}	
\usepackage{colm2024_conference}
\usepackage{hyperref}
\usepackage{url}
\usepackage{graphicx}
\usepackage{times}
\usepackage{latexsym}
\usepackage{adjustbox}
\usepackage{amsmath}
\usepackage{booktabs}
\usepackage{soul}
\usepackage{pdflscape}
\usepackage[utf8]{inputenc}
\usepackage[LFE,LAE,T1]{fontenc}
\usepackage[english,farsi]{babel}
\usepackage{pifont}
\usepackage{multirow}
\usepackage{microtype}
\usepackage{tikz}
\usepackage{algorithm}
\usepackage[noend]{algpseudocode}

\usepackage[T1]{fontenc}

\usepackage[utf8]{inputenc}

\usepackage{microtype}
\usepackage{float}
\colmfinalcopy
\makeatletter
\def\BState{\State\hskip-\ALG@thistlm}
\makeatother

\definecolor{darkblue}{rgb}{0, 0, 0.5}
\hypersetup{colorlinks=true, citecolor=darkblue, linkcolor=darkblue, urlcolor=darkblue}
\tikzstyle{border} = [draw=MidnightBlue, fill=none, very thick,
rectangle, rounded corners, inner sep=5pt, inner ysep=10pt]
\tikzstyle{title} =[draw=MidnightBlue!75, fill=MidnightBlue!20, rounded corners, text=MidnightBlue]
\newcommand{\myBox}[5]{
	\begin{tikzpicture}\node [border] (box){\begin{minipage}{0.95\textwidth}
				{#5}
		\end{minipage}};
  \node[title, left=10pt] at (box.south east) {\hbox{{\bf{#1 }}}};
  \node[title, left=55pt] at (box.south east) {\hbox{{\bf{#2 }}}}; 
  \node[title, left=110pt] at (box.south east) {\hbox{{\bf{#3 }}}};
  \node[title, left=145pt] at (box.south east) {\hbox{{\bf{#4 }}}};
  \end{tikzpicture} 
}

\tikzstyle{anotherBox} = [draw=MidnightBlue!80, fill=MidnightBlue!20, very thick,
rectangle, rounded corners, inner sep=5pt, inner ysep=10pt]
\tikzstyle{border} = [draw=MidnightBlue, fill=none, very thick,
rectangle, rounded corners, inner sep=5pt, inner ysep=10pt]

\newcommand{\sBox}[1]{
	\begin{tikzpicture}\node [anotherBox] (box){\begin{minipage}{0.8\textwidth}
				{#1}
		\end{minipage}};\end{tikzpicture} 
}

\usetikzlibrary{shapes.callouts} 

\pgfkeys{%
    /calloutquote/.cd,
    width/.code                   =  {\def\calloutquotewidth{#1}},
    position/.code                =  {\def\calloutquotepos{#1}}, 
    author/.code                  =  {\def\calloutquoteauthor{#1}},
    /calloutquote/.unknown/.code   =  {\let\searchname=\pgfkeyscurrentname
                                 \pgfkeysalso{\searchname/.try=#1,                                
    /tikz/\searchname/.retry=#1},\pgfkeysalso{\searchname/.try=#1,
                                  /pgf/\searchname/.retry=#1}}
                            }

\newcommand\calloutquote[2][]{%
       \pgfkeys{/calloutquote/.cd,
         width               = 5cm,
         position            = {(0,-1)},
         author              = {}}
  \pgfqkeys{/calloutquote}{#1}                   
  \node [rectangle callout,callout relative pointer={\calloutquotepos},text width=\calloutquotewidth,/calloutquote/.cd,
     #1] (tmpcall) at (0,0) {#2};
  \node at (tmpcall.pointer){\calloutquoteauthor};    
} 

\definecolor{darkblue}{rgb}{0, 0, 0.5}
\hypersetup{colorlinks=true, citecolor=darkblue, linkcolor=darkblue, urlcolor=darkblue}

\title{Khayyam Challenge (PersianMMLU): Is Your LLM Truly Wise to The Persian Language?}

\author{\small{Omid Ghahroodi${}^{\spadesuit\diamond}$, Marzia Nouri${}^{\spadesuit\diamond}$\thanks{These authors contributed equally to this work and are considered joint second authors. The order is listed randomly to reflect their equal contributions.} , Mohammad Vali Sanian${}^{\spadesuit\diamond*}$, Alireza Sahebi${}^{\spadesuit\diamond*}$, Doratossadat Dastgheib${}^{\spadesuit\diamond}$,}\\ \small{\textbf{Ehsaneddin Asgari${}^\ddagger$\thanks{These authors contributed equally to this work and are considered joint corresponding authors. The order of corresponding authors is listed randomly to reflect their equal contributions.} , Mahdieh Soleymani Baghshah${}^{\spadesuit\diamond\dagger}$, Mohammad Hossein Rohban}${}^{\spadesuit\diamond\dagger}$}
\\\\
$^{\spadesuit}$ Raia Center for Artificial Intelligence Research \\
$^{\diamond}$ Computer Engineering Department, Sharif University of Technology, Iran \\
$^{\ddagger}$Qatar Computing Research Institute, Qatar\\\\
\small{\texttt{\{oghahroodi98, nouri.marzia.1999, mvs2667, arsahebi97\}@gmail.com}}\\
\small{
\texttt{d\_dastgheib@sbu.ac.ir}, \texttt{easgari@hbku.edu.qa},
\texttt{\{soleymani, rohban\}@sharif.edu}
}
}

\begin{document}
\selectlanguage{english}
\maketitle
\begin{abstract}
Evaluating \textbf{L}arge \textbf{L}anguage \textbf{M}odels (\textbf{LLMs}) is challenging due to their generative nature, necessitating precise evaluation methodologies. Additionally, non-English LLM evaluation lags behind English, resulting in the absence or weakness of LLMs for many languages.
In response to this necessity, we introduce \textbf{Khayyam Challenge} (also known as \textbf{PersianMMLU}), a meticulously curated collection comprising \textbf{20,192 four-choice questions} sourced from \textbf{38 diverse tasks} extracted from Persian examinations, spanning a wide spectrum of subjects, complexities, and ages. The primary objective of the \textbf{Khayyam Challenge} is to facilitate the rigorous evaluation of LLMs that support the Persian language. Distinctive features of the \textbf{Khayyam Challenge} are \textbf{(i)} its comprehensive coverage of various topics, including literary comprehension, mathematics, sciences, logic, intelligence testing, etc aimed at assessing different facets of LLMs such as language comprehension, reasoning, and information retrieval across various educational stages, from lower primary school to upper secondary school \textbf{(ii)} its inclusion of rich metadata such as human response rates, difficulty levels, and descriptive answers \textbf{(iii)} its utilization of new data to avoid data contamination issues prevalent in existing frameworks \textbf{(iv)} its use of original, non-translated data tailored for Persian speakers, ensuring the framework is free from translation challenges and errors while encompassing cultural nuances \textbf{(v)} its inherent scalability for future data updates and evaluations without requiring special human effort. Previous works lacked an evaluation framework that combined all of these features into a single comprehensive benchmark. Furthermore, we evaluate a wide range of existing LLMs that support the Persian language, with statistical analyses and interpretations of their outputs. We believe that the \textbf{Khayyam Challenge} will improve advancements in LLMs for the Persian language by highlighting the existing limitations of current models, while also enhancing the precision and depth of evaluations on LLMs, even within the English language context.
\end{abstract}

\section{Introduction}

\textbf{L}arge \textbf{L}anguage \textbf{M}odels (\textbf{LLMs}) have recently revolutionized applications of machine intelligence \citep{hong20243d, wu2023autogen, thirunavukarasu2023large, glukhov2023llm}. The rapid deployment of these models within industrial and public sector solutions has made evaluating their capabilities an imperative task \citep{guo2023evaluating}.
To address critical aspects of language understanding and bridge the gap between the knowledge that models observed during pretraining and the measures of success, the MMLU benchmark was introduced \citep{hendryckstest2021}. It assesses the ability of LLMs across a diverse set containing 57 subjects.

Although automatic translation efforts like MMLU can bridge the gap in evaluation resources for languages other than English \citep{achiam2023gpt}, the cultural specificity of these questions limits their applicability across different languages and cultures. This is particularly true for Persian, a language with its own rich culture and literature, where direct translation of English MMLU content may not be suitable for certain culture-specific subjects. In addition, our investigation reveals significant complications in automatic translation, particularly in specialized domains like Physics, which can be attributed to the quality of state-of-the-art translations for specific subjects. Despite advancements in machine translation technology, accurately conveying precise meaning in certain domains remains a hard task. For instance, the GPT-3.5 automatic translation often fails to include essential units such as
"microjoules" in translations, leading to inaccuracies. Also, loss of information can occur in translation due to the existence of Persian words that lack exact equivalents in English. For example, in Persian, there are distinct terms for paternal aunt, maternal aunt, paternal uncle, and maternal uncle, while in English, only ``aunt" and ``uncle" are used (See Table \ref{table:trans_sample}). This underscores the inherent difficulty in translating domain-specific terminology accurately. For such reasons, some studies have evaluated LLMs on original non-English datasets rather than the translated ones. For instance, \cite{li2023cmmlu} has introduced a Chinese dataset across 67 topics and indicated that current models struggle to achieve accuracies above a certain threshold. Also,  \cite{zhang2023mexam} introduced a multilingual and multimodal dataset, and showed that multilingual text processing hardly achieves over $60\%$ accuracy.



Motivated by the mentioned issues, we propose the Khayyam Challenge, also referred to as PersianMMLU, a benchmark designed to analyze the performance of LLMs in Persian and evaluate their knowledge and abilities comprehensively.
Named in honor of the famed Persian polymath Omar Khayyam, whose contributions spanned various disciplines including Mathematics, Astronomy, Philosophy, and Poetry, the Khayyam Challenge embodies the multidimensional nature of Persian language understanding.
This benchmark covers 38 subjects, including Mathematics and Physics, which require reasoning and computational ability, to humanities and social sciences, demanding nuanced understanding and cultural sensitivity. Unlike the previous Persian datasets such as ParsiNLU \citep{khashabi2021parsinlu}, our benchmark includes more diverse topics in addition to different educational stages. 
Moreover, our dataset distinguishes itself from previous efforts like ParSQuAD \citep{Abadani2021parsquad} by being originally constructed in Persian, naturally incorporating the nuanced semantics and intricacies inherent to the language itself, rather than solely focusing on literal translations.
Unlike PersianQA \citep{PersianQA} and PQuAD \citep{DARVISHI2023101486}, which are extractive datasets where models are tasked with extracting answers from given paragraphs and questions, our benchmark offers a more comprehensive evaluation of LLMs. This is because the task of answer extraction alone may not sufficiently assess the models' overall language understanding and reasoning capabilities.
Our proposed dataset contains ``Iran's national university entrance", and Kanoon Farhangi Amoozesh (Cultural Educational Center), wherein questions are not only verified and validated by experts in each subject but also accompanied by metadata for each question. This metadata includes the difficulty level, a descriptive answer, educational stage, subject, and the specific topic of the question. Through the Khayyam Challenge, we aim to provide a holistic evaluation framework that reflects the diverse linguistic and cognitive challenges inherent in processing Persian text across various domains. 

\begin{table*}
\begin{center}
\scalebox{0.68}{
\begin{tabular}{|c|c|}
\hline

\multirow{2}{*}{\textbf{question}}  \rule{0pt}{2.5ex}       & \begin{otherlanguage}{farsi} \textRL{
شنونده‌ای که مساحت پرده گوشش $60$ میلی‌متر مربع است، تراز شدت صوت حاصل از یک منبع را $50$ دسی‌بل ،
}\end{otherlanguage}\\  
                                  & \begin{otherlanguage}{farsi} \textRL{
احساس می‌کند انرژی که در مدت $50$ ثانیه به پرده گوش این شنونده می‌رسد، چند میکروژول است؟}\end{otherlanguage} \rule[-1.5ex]{0pt}{0pt} \\ \hline
\multirow{2}{*}{\textbf{correct translation}}  \rule{0pt}{2.5ex}                         & A listener with an eardrum area of $60 mm^2$ perceives the sound intensity level from a source as $50$ decibels. \\ 
                       & How much energy reaches the listener's ear in $50$ seconds in microjoules? \rule[-1.5ex]{0pt}{0pt}\\ \hline
\multirow{2}{*}{\textbf{API translation}}  \rule{0pt}{2.5ex}                              & A listener with an ear canal area of $60 mm^2$ perceives the sound intensity level from a source as $50$ decibels. \\ 
 & How much energy reaches the listener's ear in 50 seconds? \rule[-1.5ex]{0pt}{0pt}\\ \hline
\hline
\multirow{2}{*}{\textbf{question}}  \rule{0pt}{2.5ex}       & \begin{otherlanguage}{farsi} \textRL{در بیت زیر، شاعر از کدام اختیارات شاعری بهره برده است؟ }\end{otherlanguage}  \\  
                                  & \begin{otherlanguage}{farsi} \textRL{ »گویی بطّ سفید جامه به صابون زده است / کبک دری ساق پای در قدح خون زده است«}\end{otherlanguage} \rule[-1.5ex]{0pt}{0pt} \\ \hline
\multirow{3}{*}{\textbf{correct translation}}  \rule{0pt}{2.5ex}                         &  In the following verse, which poetic devices has the poet employed?\\ 
                       & ``The duck's feathers are so white, as if they've been washed with soap / \\
& 
The partridge's legs are so red, as if they've been dipped in a cup of blood" \rule[-1.5ex]{0pt}{0pt}\\ \hline

\multirow{2}{*}{\textbf{API translation}}  \rule{0pt}{2.5ex}                             & In the following verse, which poetic options did the poet use?\\ 
 & ``It's as if a white duck has put its clothes on soap / the partridge has put its leg in a glass of blood" \rule[-1.5ex]{0pt}{0pt} \\ \hline
\hline
\multirow{2}{*}{\textbf{question}}  \rule{0pt}{2.5ex}       & \begin{otherlanguage}{farsi} \textRL{ کیان پسر عمه زهرا، پسر خاله من است. من $2$ خاله و $3$ دایی دارم.
 }\end{otherlanguage}  \\  
                                  & \begin{otherlanguage}{farsi} \textRL{ زهرا به‌ترتیب چند عمه و چند عمو دارد؟ }\end{otherlanguage}  \\ \hline
\multirow{2}{*}{\textbf{correct translation}} \rule{0pt}{2.5ex}                          & Kian is the son of the sister of Zahra's Father, and he is the son of my mother's sister. \\                        &  My mother has two sisters and three brothers. How many sisters and brothers does Zahra's father have? \rule[-1.5ex]{0pt}{0pt} \\ \hline
\multirow{2}{*}{\textbf{API translation}} \rule{0pt}{2.5ex}                              & Kian is the son of my aunt Zahra. I have 2 aunts and 3 uncles.  \\ 
 & How many aunts and how many uncles does Zahra have respectively? \rule[-1.5ex]{0pt}{0pt} \\ \hline
\end{tabular}
}
\end{center}
\caption{Examples illustrating information loss in domain-specific translations from English to Persian Using the GPT-3.5 API.}
\label{table:trans_sample}
\end{table*}  

In our evaluations, we assessed several state-of-the-art language models, including GPT-3.5, GPT-4 \citep{openai2023gpt4}, Aya \citep{ustun2024aya}, PersianMind \citep{persianmind}, mGPT \citep{mGPT}, mT0 \citep{muennighoff2022crosslingual}, Claude3-haiku \citep{Claude3}, and XVERSE\footnote{https://github.com/xverse-ai}, all purportedly equipped with some level of understanding of the Persian language. Our findings indicate that while most of these models struggle to grasp Persian nuances, particularly evident in domains such as Calculus, Logic, and Geometry where accurate comprehension is essential, some exhibit comparatively better performance in contexts reliant on contextual understanding, such as Economics, Psychology, and Social studies. Notably, GPT-4 showcased relatively improved performance across multiple domains. However, there remains a clear imperative need for further enhancements across all models, especially in technical disciplines like Discrete Mathematics, where precise language comprehension is paramount for meaningful outcomes.

Furthermore, our evaluation unveiled new insights. We observed dependencies between the perceived difficulty of questions for humans versus LLMs. While both encounter similar challenges in tackling difficult questions, LLMs like GPT-4 demonstrated superior accuracy compared to humans, in questions intentionally designed to deceive, referred to as trapped questions.
Moreover, our analysis identified biases in certain language models' responses, such as GPT-3.5 favoring particular choices,  suggesting room for further improvement in these models.

Overall, the Khayyam Challenge (PersianMMLU) marks a significant step forward in evaluating the language understanding and abilities of LLMs that support
the Persian language.

\section{Related work}
\subsection{Large Language Models}
Over the past few years, there has been a significant improvement in the performance of language models. This progress has been observed in line with the scaling law \citep{kaplan2020scaling}, thanks to the increasing size of training datasets, enhanced processing power, and new evolved model architectures. The continuing process of scaling the models resulted in LLMs like GPT-3 \citep{Brown2020gpt3},
GPT-4 \citep{openai2023gpt4},
Claude3,
mT0\citep{muennighoff2022crosslingual},
XVERSE, 
Aya \citep{ustun2024aya},
etc. 

Even though AI models are highly capable of solving various tasks, they continue to encounter difficulties when it comes to real-world problems that, for example, require strong reasoning abilities or complex mathematical calculations \citep{Chang2024survey, zhong2023agieval}. Therefore, we need to assess the effectiveness of these models in solving high-level tasks. This enables us to identify the weak points of the models and work towards improving them in the future.

Despite some of the recent LLMs being multilingual, studies indicate that their effectiveness is not as pronounced in non-Latin or low-resource languages as it is in English \citep{zhang2023mexam}. Consequently, it is essential to assess multilingual LLMs on tasks that employ languages other than English.

\subsection{Evaluation of LLMs}

Several benchmarks have been developed to assess the performance of LLMs. One of the most significant benchmarks is MMLU \citep{hendryckstest2021}, which evaluates language models for answering multiple-choice questions in 57 different tasks, but only in English. M3Exam \citep{zhang2023mexam} introduces a multilingual, multimodal, and multilevel benchmark for evaluating LLMs including more than 12K multiple-choice questions from 9 languages (excluding Persian) at three educational stages. AGIEval \citep{zhong2023agieval} is another benchmark that assesses the performance of LLMs on human-centric standardized exams in English and Chinese languages to measure their ability in human-level tasks.

There have been a few benchmarks built to assess language models on the Persian language, including ParSQUAD \citep{Abadani2021parsquad}, PersianQA \citep{PersianQA}, ParsiNLU \citep{khashabi2021parsinlu}, and PQuAD \citep{DARVISHI2023101486}, in which some of their features are compared in Table \ref{tab:otherdatasets}. ParSQUAD, PersianQA, and PQuAD present extractive datasets where models are asked to extract answers from given paragraphs. While this task can gauge models' reading comprehension skills, it may not effectively evaluate general capabilities and inherent knowledge of models. ParsiNLU evaluates language models based on 14500 questions from six language understanding tasks, including multiple-choice QA (MCQA), sentiment analysis, and more (Table \ref{tab:otherdatasets}). 
However, the questions in this benchmark fail to reach human-level complexity, thus inadequately assessing important skills of LLMs, including complex reasoning, needed for solving higher educational stages questions.
Moreover, ParsiNLU covers only three subject tasks in MCQA and lacks adequate metadata, such as question difficulty levels. This deficiency further restricts our capacity to evaluate the model's proficiency across specific tasks and different difficulty levels.

To address this gap, we introduce Khayyam Challenge, which features multiple-choice questions sourced from high-standard exams, and contains reach metadata to evaluate the innate knowledge and human-like skills of LLMs across different difficulty levels and educational stages.

\begin{table*}
\centering
\scalebox{0.60}{
\begin{tabular}{|c||c|c|c|c|c|c|c|c|}
    \hline
    \multirow{2}{*}{\textbf{Benchmark}} & \multirow{2}{*}{\textbf{Languages}} & \multirow{2}{*}{\textbf{Type}} & \multirow{2}{*}{\textbf{NLU Tasks}} & \multirow{2}{*}{\textbf{\# Instance}} & \multicolumn{4}{|c|}{\textbf{Metadata}} \\
    \cline{6-9}
     &  &  &  &  & \textbf{Desc. Ans.} & \textbf{Diff. Lev.} &  \textbf{Trap} & \textbf{\# Task} \\
    \hline
    \hline
    MMLU & Eng. & orig. & MCQA & 15,908 & \ding{53} & \ding{53} & \ding{53} & 57 \\
    \hline
    AGIEval & Eng., Chi. & orig.+trans. & MCQA, FIB & 8,062 & some instances & some instances & \ding{53} & 20 \\
    \hline
    M3Exam & 9 (no Persian) & orig. &  MCQA  & 12,317 & \ding{53} & = edu. stages & \ding{53} & 4\\
    \hline
    ParSQUAD & Persian & trans. & RC & 70,560 & - & \ding{53} & - & 477 titles \\
    \hline
    PersianQA & Persian & orig. & RC & 9,938 & - & \ding{53} & - & 991 titles \\
    \hline
    PQuAD & Persian & orig & RC & 80,000 & - & \ding{53} & - & 19 \\
    \hline
    ParsiNLU & Persian & orig.+trans. & MCQA, RC, SA, TE, QP, MT & 14,500 & \ding{53} & \ding{53} & \ding{53} & 3 in MCQA \\
    \hline
    Khayyam Challenge & Persian & orig. & MCQA & 20,192 & \ding{51} & \ding{51} & \ding{51} & 38 \\
    \hline
\end{tabular}}

\caption{Comparison of various features of Persian and English benchmarks. As descriptive answers and trapped questions were not defined for reading comprehension benchmarks, we marked those fields with a hyphen (-).  Desc. Ans.: Descriptive Answer, Diff. Lev.: Difficulty Level, Eng: English, Chi: Chinese, orig.: original non-translated question, tran.: translated question, MCQA: Multiple-Choice Question Answering, FB: Fill in the Blank, RC: Reading Comprehension, SA: Sentiment Analysis, TE: Textual Entailment, QP: Question Paraphrasing, MT: Machine Translation., edu. stages: educational stages}
\label{tab:otherdatasets}
\end{table*}


\section{Data}
The Khayyam Challenge presents a robust dataset aimed at enhancing the evaluation of LLMs that support Persian, particularly in the context of multiple-choice question answering. This dataset encompasses a diverse range of subjects, reflecting a comprehensive approach to assessing various cognitive abilities including language comprehension, reasoning, and knowledge recall across different educational stages.
The educational system in Iran, from which this dataset draws, is structured into 12 years of schooling divided into segments: 6 years of primary education and 6 years of secondary education. Primary education is split into lower primary school (LPS) for the first 3 years, followed by upper primary school (UPS) for the next 3 years. Secondary education is similarly divided, with lower secondary school (LSS) encompassing the first 3 years, and upper secondary school (USS) comprising the final 3 years.
\subsection{Data construction}
The dataset originates from the ``Pellekan Yadgiri (Learning Ladder)'' website\footnote{\href{https://peleyad.com/}{https://peleyad.com/}}, a part of the Kanoon Farhangi Amoozesh\footnote{\href{https://en.kanoon.ir/}{https://en.kanoon.ir/}} (Cultural Educational Center), a renowned private educational institution in Iran. Since conducting its first examination in 1993, Kanoon has been at the forefront of educational innovation, serving a nationwide network of 450,000 students. The center is highly regarded for its comprehensive range of educational services, notably its facilitation of creating and administering smart, customized tests and providing standardized solutions for exercises through the Pellekan Yadgiri platform. This initiative aims to enable educators to design and administer quality, customized tests effortlessly and to provide students with insightful feedback to enhance their learning process.

The quality of the questions in the dataset is exceptionally high, a testament to the institution's reputable standing in the educational sector. The ongoing development of new questions for various educational stages and subjects allows for the continuous expansion and updating of the dataset, reducing the risks of data contamination and erosion. A subset of the questions also includes items from the national university entrance examination in Iran, adding to the dataset's authenticity and relevance.

\subsection{Metadata characteristics}
The Khayyam Challenge is enriched with valuable metadata that elevates its utility beyond a simple aggregation of questions. This metadata includes:

\begin{figure}
    \centering
    \begin{tabular}{cc}
     \includegraphics[width=5cm]{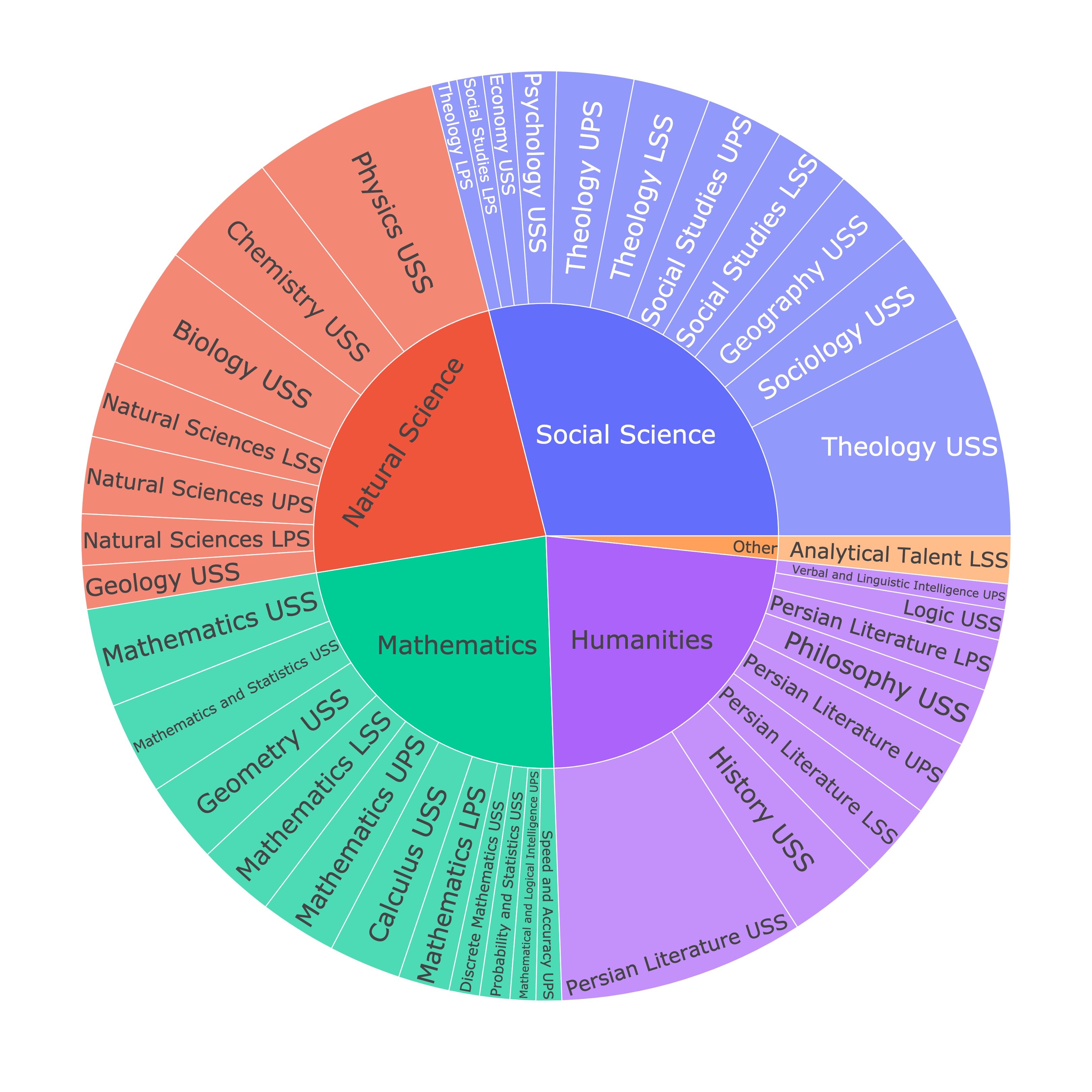} & \includegraphics[width=5cm]{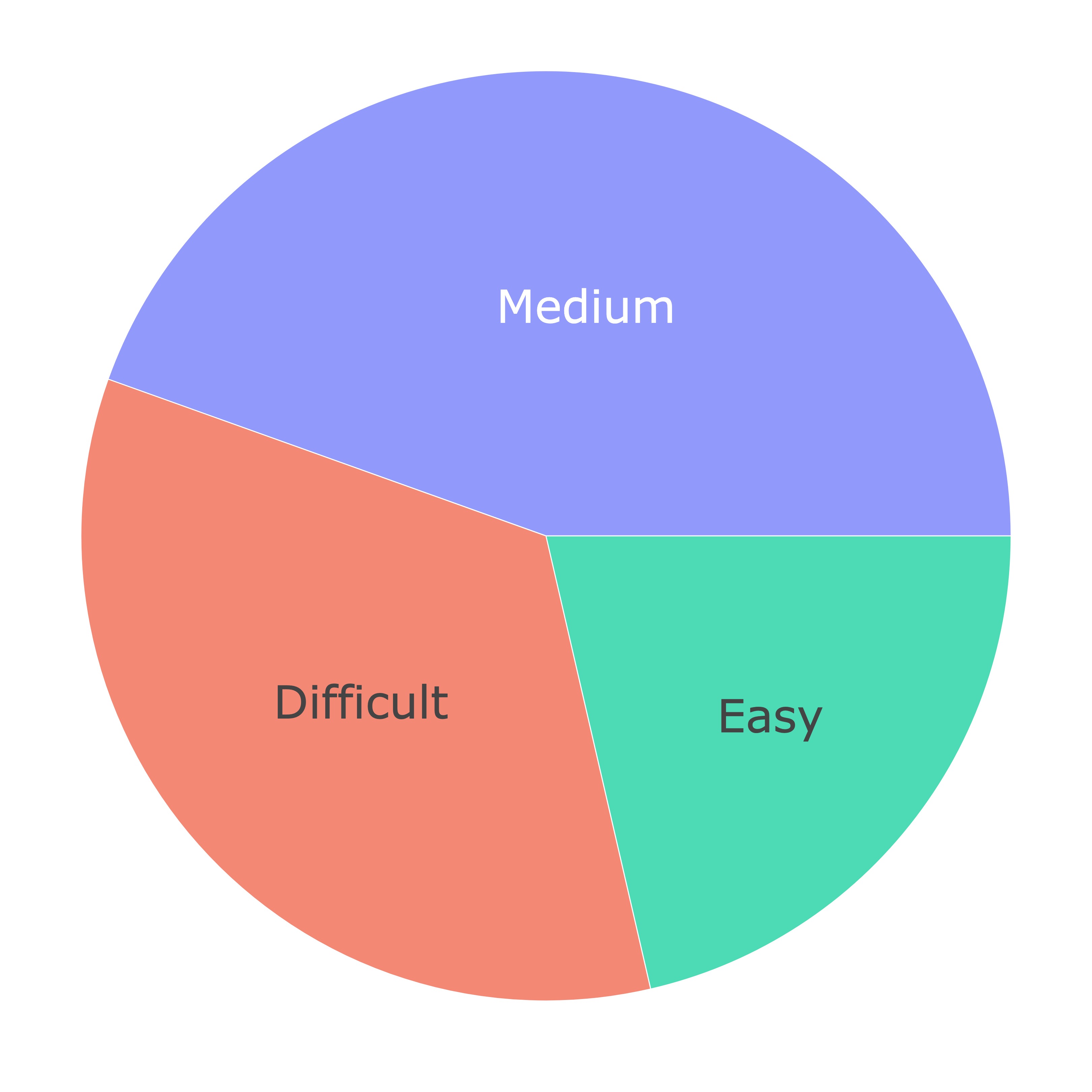} \\
     (a) & (b)
    \end{tabular}
    \caption{Question distribution with respect to categories, level of difficulty, and educational stages. LPS: Lower Primary School, UPS: Upper Primary School, LSS: Lower Secondary School, USS: Upper Secondary School}
    \label{fig:question_distribution_educational_stages_And_difficulty}
\end{figure}

\begin{itemize}
\item \textbf{Educational Stage}: Specifies the educational stage for which the question is intended (LPS, UPS, LSS, USS), allowing for the assessment of appropriateness and difficulty relative to the expected knowledge base at each stage of education.
    \item \textbf{Difficulty Level}: Each question is classified into one of five distinct difficulty levels: easy,  relatively easy, medium, relatively difficult, and difficult. This nuanced categorization allows a detailed analysis of question difficulty and examinee performance.
    \item \textbf{Descriptive Answers}: In addition to the correct answer, our dataset provides a detailed explanation for each question. This is crucial for understanding the reasoning behind the correct answer, facilitating a deeper comprehension of the question.
    \item \textbf{Trap}: Human experts have identified if a question contains a "trap" choice—an incorrect answer that might be easily mistaken for the correct one. These questions are referred to as "trapped questions" and are generally more challenging, with the majority classified as difficult. This helps in understanding common misconceptions and the effectiveness of question design in truly testing knowledge and reasoning abilities.
    \item \textbf{Human Performance}: This metric quantifies the percentage of students answering a question correctly.
    \item \textbf{Specific Topic}: Questions are meticulously categorized into detailed subjects, such as "Mathematics > Discrete Maths > Combinatorics." This detailed classification enables targeted analysis of exam content and provides insights into the distribution and depth of questions across various subjects.
    \item \textbf{Year}: Indicates the year when the question was designed, which can provide insights into the evolution of question complexity and educational standards over time.
    
\end{itemize}

The inclusion of this metadata is not merely for augmentative purposes; it serves a critical role in enabling comprehensive analyses that can benefit educators, researchers, and developers of educational technologies. Specifically, it allows for the comparison of performance between human examinees and LLMs on specific topics under varying difficulty levels. By assessing whether LLMs fall for the traps as humans or how they approach questions requiring complex thought processes, we can gain valuable insights into the capabilities and limitations of current AI technologies in educational contexts.

Moreover, the presence of descriptive answers supports the development of more sophisticated AI models by facilitating "chain of thought" processing, where the model learns to approach a problem step-by-step, mirroring human problem-solving methods. This not only enhances the model's problem-solving skills but also its ability to explain its reasoning in a manner that is understandable to humans.

\subsection{Data statistics}
The dataset contains 20,192 multiple-choice questions across 38 tasks, spanning subject areas like humanities, mathematics, natural science, and social science, along with elements of intelligence testing. These questions necessitate a blend of knowledge and reasoning. Additionally, the dataset includes 15,933 questions with human performance data, excluding Iran's national university entrance exam questions, and features 3,531 trapped questions. Figure \ref{fig:question_distribution_educational_stages_And_difficulty}-(a) depicts the allocation of questions among the main categories and their respective tasks. Figure \ref{fig:question_distribution_educational_stages_And_difficulty}-(b) outlines the distribution of questions based on their levels of difficulty. For more detailed information about the data, refer to Appendix \ref{sec:data}.

\subsection{Key features}
The Khayyam Challenge Dataset stands out for several reasons:
\begin{itemize}
\item \textbf{Comprehensive Coverage}: It spans a broad spectrum of subjects from literary comprehension to logic and intelligence testing, catering to different stages of education. This diversity makes it a versatile tool for assessing language models' capabilities across various domains.
\item \textbf{Rich Metadata}: The inclusion of detailed question metadata enhances the dataset's utility for nuanced analysis and model evaluation, providing valuable context for each question.
\item \textbf{New Data Utilization}: By incorporating questions never before used in research, the dataset avoids common data contamination issues, offering a fresh challenge to language models.
\item \textbf{Original, Non-Translated Content}: Focused on the Persian language, the dataset eliminates translation errors and incorporates cultural nuances, making it uniquely valuable for related linguistic and cultural studies.
\item \textbf{Scalability}: The dataset's design and sourcing methodology ensure its adaptability and expandability, allowing for straightforward updates and extensions without substantial human intervention.

\end{itemize}

\section{Experiments}

\begin{figure}[h]
\begin{center}
\includegraphics[width=0.5\linewidth]{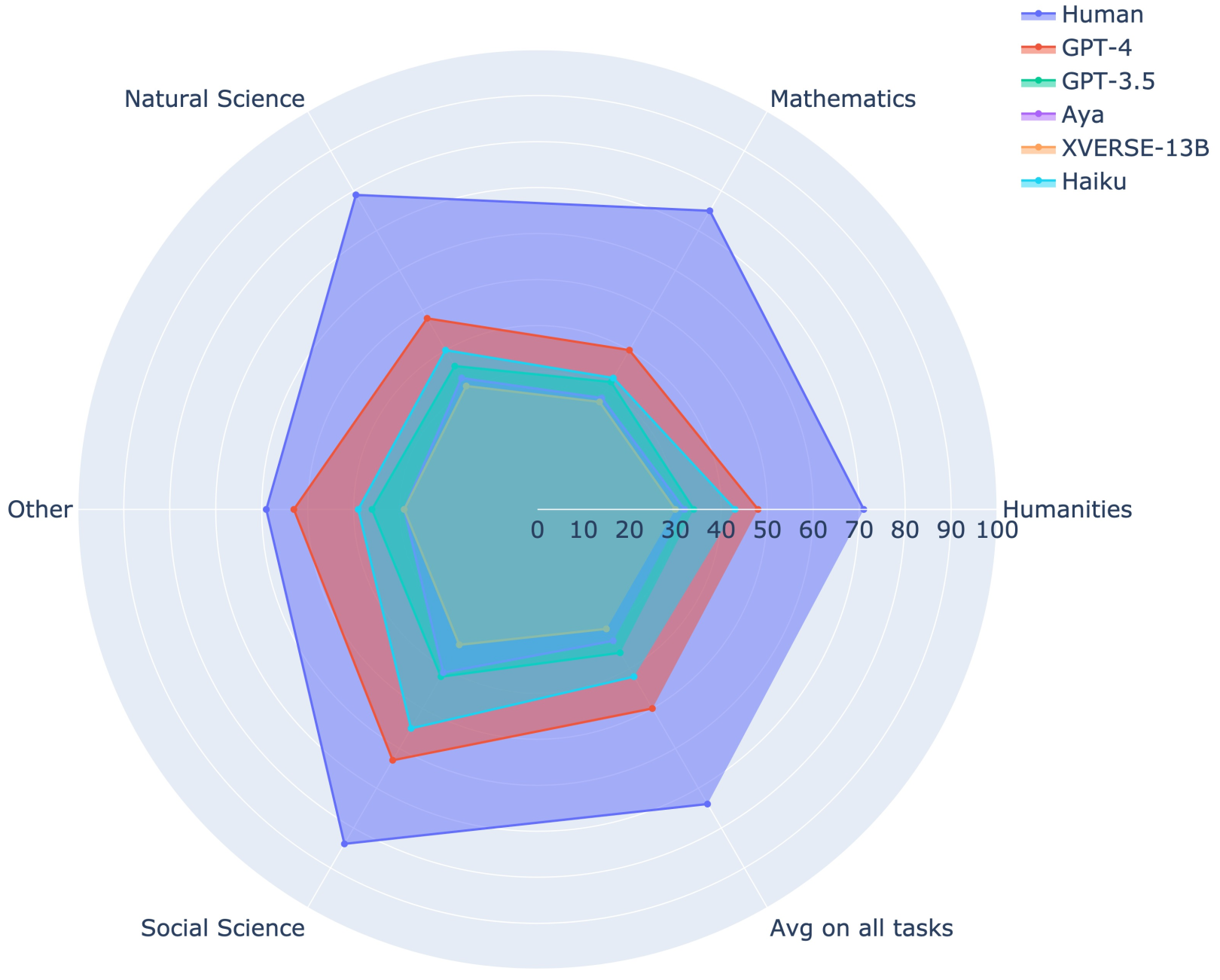}
\end{center}
\caption{Comparison of accuracy across main categories for humans and various models}
\label{fig:radarDiag}
\end{figure}




\subsection{Settings}

We evaluated our benchmark using nine distinct models: GPT-4, GPT-3.5, Aya, XVERSE-13B, XVERSE-7B, PersianMind, mT0XL, mGPT, and Claude 3/Haiku. Detailed descriptions of each model can be found in Appendix \ref{sec:models}. Additionally, we benchmarked a standardized prompt-0 \ref{fig:prompt0} template on the entire dataset and two other templates, prompt-1 \ref{fig:prompt1} and \ref{fig:prompt2}, on a subset of 1000 samples from the dataset. We also conducted Chain-of-Thought (CoT) \ref{fig:prompt-cot} on a subset of 1000 samples requiring CoT, such as mathematical questions. To ensure consistency and fairness throughout our experiments, we kept all model hyperparameters at their default values. Furthermore, we set the temperature parameter to zero for all models and did not impose any maximum limit on the number of tokens, allowing models the freedom to conduct any type of inference they desired.

\subsection{Answer extraction methods}

We employed three distinct techniques to extract answers: Regex, Single Token Probability, and Full Answer Probability, and conducted an evaluation of their individual accuracies. In the Regex method, we developed detailed Regex functions tailored to each model, which is available in our framework, to accurately capture the desired choices. In cases where Regex failed to identify the correct option, we utilized a pre-trained model to generate embeddings for each choice within the question, selecting the most similar one as the correct response. Furthermore, in our analysis, we considered human accuracy by examining our comprehensive metadata containing response percentages for each question choice. We chose the answer selected by humans if the combined total of the remaining three options was lower.

In the Single Token Probability approach, applied to models providing token probabilities, we calculated the softmax probability of tokens and selected the most probable tokens from 1, 2, 3, and 4 in Persian as the response. As for the Full Answer Probability method as we can see at formula \ref{eq:answer_equation}, we tokenized the text of each choice, computed the sum of the logarithm of token probabilities for each option, standardized it to the length of each option, and chose the most likely option as the answer.
\begin{equation}
\text{Answer} = \arg\max_i \left( \frac{1}{n_i} \sum_{j=1}^{n_i} \log(p_{ij}) \right)
\label{eq:answer_equation}
\end{equation}

Our analysis shows that the Regex method has the highest accuracy, whereas the full prob method demonstrated a significant decrease in accuracy compared to the other two approaches.

\subsection{Impact of translation quality}

We aim to assess the impact of translation from state-of-the-art (SOTA) translation models on performance. To do so, we selected a set of sample questions from the dataset and translated them with the assistance of field experts, as well as using off-the-shelf translation models. Following this, we evaluated the models' performance on both sets of samples: those translated with expert assistance and those translated by models alone. Our findings revealed a notable decrease in performance when translated with the help of off-the-shelf models, highlighting the need for a new dataset that does not depend solely on translated data.

\subsection{Limitations of few-shot approach}
     
  We have developed a benchmark code for our dataset and used the zero-shot and CoT methods to calculate accuracy. Previous studies \citep{li2023cmmlu, zhong2023agieval, zhang2023mexam} shows that using few shots on instruction-tuned models does not enhance accuracy and may even may decrease it. Therefore, we did not measure accuracy using few-shot techniques.

\begin{table}[H]
\begin{center}
\scalebox{0.6}{
    \begin{tabular}{|c||c|ccccccccc||c|}
    
        \hline
        \textbf{Main categories} & \textbf{Human} & \textbf{GPT-4} & \textbf{Haiku} & \textbf{GPT-3.5} & \textbf{Aya} & \textbf{XVERSE-13B} & \textbf{XVERSE-7B} & \textbf{PersianMind} & \textbf{mT0XL} & \textbf{mGPT} & \textbf{Random} \\ \hline \hline
        Humanities & 0.71 & 0.48 & 0.43 & 0.34 & 0.32 & 0.3 & 0.28 & 0.28 & 0.28 & 0.26 & 0.25 \\ \hline
        Mathematics & 0.75 & 0.4 & 0.33 & 0.32 & 0.28 & 0.27 & 0.26 & 0.26 & 0.26 & 0.26 & 0.25 \\ \hline
        Natural Science & 0.79 & 0.48 & 0.4 & 0.36 & 0.33 & 0.31 & 0.28 & 0.29 & 0.28 & 0.25 & 0.25 \\ \hline
        Social Science & 0.84 & 0.63 & 0.55 & 0.42 & 0.41 & 0.34 & 0.32 & 0.3 & 0.32 & 0.25 & 0.26 \\ \hline
        Other & 0.59 & 0.53 & 0.39 & 0.36 & 0.29 & 0.29 & 0.32 & 0.27 & 0.28 & 0.28 & 0.26 \\ \hline \hline
        Avg on all tasks & 0.74 & 0.5 & 0.42 & 0.36 & 0.33 & 0.3 & 0.29 & 0.28 & 0.28 & 0.26 & 0.25 \\ \hline
        Avg on all questions & 0.77 & 0.5 & 0.43 & 0.36 & 0.34 & 0.31 & 0.29 & 0.28 & 0.29 & 0.25 & 0.25 \\ \hline
        

    \end{tabular}}
\end{center}
\caption{Human vs model accuracy for main categories: model answers extracted via Regex}  
\label{table:acc_main_catgories_regex}
\end{table}

\section{Results and discussions}

\subsection{Results across all models}
The evaluation results of zero-shot method for all models across five main categories and three choice-extraction methods (regex+, single token probability, and full answer probability), as well as human performance, are presented in the Table \ref{table:acc_main_catgories_regex}, and appendix Tables \ref{table:acc_main_catgories_probability_approach}, \ref{table:acc_main_catgories_FullProbability_approach}. The more comprehensive results on all 38 tasks are reported in the appendix (Tables \ref{table:Acc_regex_all_subjects}, \ref{table:Acc_prob_all_subjects}, \ref{table:Acc_fullprob_all_subjects}). Also the results of CoT and its comparison with zero-shot for GPT-3.5 using regex+ method across three main categories on a subset of dataset with 1000 questions are presented in appendix Table \ref{table:cot}. These results yield the following key findings:
\begin{itemize}

    \item Utilizing Regex method for answer extraction (Table \ref{table:acc_main_catgories_regex}) results in highest model performance compared to Single Token Probability (Table \ref{table:acc_main_catgories_probability_approach}) and Full Answer Probability (Table \ref{table:acc_main_catgories_FullProbability_approach}) due to its more accurate and comprehensive choice extraction procedure. In the rest of the results section, we compare the accuracy of models using the Regex method. 
    
    \item GPT-4 outperforms all other models in all five main categories, with an average accuracy of 8 percent higher than Claude3-haiku, the second-best performing model.
    
    \item Aya, an open-source model, performs comparably or even better than GPT-3.5, a closed-source model, in 8 tasks including Sociology USS and Theology LSS. This demonstrates the convergence of open-source models' capabilities towards closed-source models.
    
    \item Although PersianMind, a 7B Persian-English LLM, is trained and fine-tuned on 2 billion Persian tokens \citep{persianmind}, its performance is weaker than mT0XL, a multilingual 3.7B LLM.
    
    \item The performance gap between the best-performing model, GPT-4, and human averages around 35\%. In subjects like mathematics, this gap widens to 50\% accuracy for GPT-4. This exhibits a real challenge of current LLMs in solving human-level questions, especially in complex mathematical questions that need high-level mathematical calculation and reasoning skills.

    \item The models exhibit weaker performance in mathematics and natural science main categories compared to humanities and social sciences. This indicates their weaker performance on questions requiring high reasoning skills, compared to those mainly reliant on models' inherent knowledge. This underscores the necessity for enhancing the models' reasoning ability in the Persian language.

    \item The CoT has improved the performance of GPT-3.5 in Mathematics questions by 10\%, but decreased the performance in Humanities questions by 8\%.
\end{itemize}

\subsection{Accuracy trends}
We demonstrated the accuracy of the models on all three difficulty levels of questions for different educational stages and question publication years in Figure \ref{fig:question_dist_educational_stage} and Figure \ref{fig:question_dist_published_year} in the Appendix \ref{sec:accuracy_candidate_options}, respectively. Key findings from these figures include:
\begin{itemize}
    
    \item Most model accuracies decline with the increasing publication year for questions with medium and easy difficulty levels, while human performance remains consistent. This suggests that humans may have adapted to the evolving questions' difficulty over time, whereas models have not.

    
    \item There exists a notable performance gap between difficult and medium questions in human performance, indicating that difficulty level has a stronger impact on human performance compared to LLMs. Since humans assigned difficulty levels to questions, this gap may stem from differing perceptions of difficulty between humans and LLMs.
    
    \item GPT-4 outperforms humans in difficult questions within four years, as well as during the initial two educational stages (LPS and UPS). This result suggests that modern LLMs may excel at analyzing difficult questions compared to humans.
\end{itemize}

\subsection{Selected choice distribution}
Figures \ref{fig:Choice_dist_regex}, \ref{fig:Choice_dist_sing_prob}, \ref{fig:Choic_dist_fullProb} in the Appendix \ref{sec:accuracy_candidate_options} depicts the selected choice distributions of various models. It reveals that despite GPT-3.5 having the lowest count of unanswered questions, GPT-4 surpasses it in accuracy, indicating a more refined understanding despite its higher non-response rate. Comparatively, GPT-4 exhibits a more uniform distribution of choice selection, aligning closely with the Ground Truth and showing less bias than its predecessor, GPT-3.5, which tends towards selecting the second and third choices. High-performance models like GPT-4 demonstrate a closer alignment with Ground Truth distribution, indicating a lower bias level and potentially higher utility in applications.

\subsection{Trap analysis}
Table \ref{table:trap} in the Appendix \ref{sec:trap} compares the performance of models and humans on trapped questions, using an x/y format where x represents overall accuracy and y the accuracy on trapped questions. The data shows that while traps often mislead students, leading to nearly random performance, models like GPT-4 exhibit only a slight drop in accuracy when faced with these traps. However, this drop in accuracy is most pronounced in the Social Sciences and Humanities. Notably, GPT-4 outperforms humans across all main categories in handling trapped questions, indicating its robustness against misleading choices and affirming the different perspectives of difficulty between humans and AI models.


\subsection{Difficulty levels analysis}
To explore how humans and LLMs perceive the difficulty of questions, we examined the accuracy of different models across three difficulty levels. Although our dataset included five labels ranging from easy to difficult to denote question difficulty, we combined the two relatively easy and relatively difficult labels into ``easy" and ``difficult", respectively, to ensure a more balanced distribution of question difficulty. Our experiment revealed a consistent trend: as question difficulty increased, both human and model answering accuracy decreased (See Tables \ref{table:acc_over_different_diff_lvls_reg}, \ref{table:acc_over_different_diff_lvls_single_probab}, \ref{table:acc_over_different_diff_lvls_full_probab} in the Appendix \ref{sec:accuracy_candidate_options}).

Notably, in analytical and knowledge-based topics such as the humanities category, the GPT-4 model demonstrated superior performance compared to humans in tackling difficult questions. Conversely, models exhibiting more human-like performance, such as GPT-4, revealed that humans significantly outperformed them on easier questions.
For additional statistical results, refer to Appendix \ref{sec:statistical_tests}.

\section{Conclusions}

We introduced \textbf{Khayyam Challenge}, also known as \textbf{PersianMMLU}, as the first framework for assessing LLMs in the Persian language across various tasks, difficulty levels, and educational stages. This framework includes comprehensive metadata such as human performance, difficulty levels, and traps. Our assessment encompassed examining the performance of current LLMs on these datasets, evaluating their ability to extract answers (probabilistic and rule-based paradigms), and considering various aspects highlighted in the metadata. Our findings revealed that while LLMs demonstrated relatively satisfactory performance in question-solving tasks (especially on GPT-4), they still significantly lagged behind human performance, particularly in tasks necessitating reasoning. Moreover, analysis of metadata concerning difficulty levels and trapped questions unveiled notable discrepancies between model and human behavior, suggesting fundamental differences in learning approaches. This underscores the necessity for adaptations in LLMs training methodologies to achieve human-like proficiency. For future works, we aim to develop an LLM that bridges the performance gap between existing open-source models and GPT-4.

\section{Data Restriction}
This dataset is distributed under a Creative Commons No Derivatives (CC ND) license, prohibiting the creation of derivative works. It is designated exclusively for non-commercial, academic research to prevent conflicts with Kanoon Farhangi Amoozesh's (Cultural Educational Center) educational activities. Researchers are obligated to adhere to this condition, ensuring their utilization of the dataset remains confined to academic research purposes and respects the restrictions of the CC ND license.

\section{Acknowledgement}

The authors would like to express their sincere gratitude to Pellekan Yadgiri (Learning Ladder), a sub-organization of Kanoon Farhangi Amoozesh (Cultural Educational Center), for providing the valuable dataset and related metadata that was crucial for this research. We are particularly grateful to Masoud Tajfard for his kind assistance in facilitating the acquisition of this data, which was permitted for research purposes. The dataset and metadata made available by Pellekan Yadgiri were instrumental in enabling the analyses and findings presented in this work.

\bibliography{colm2024_conference}
\bibliographystyle{colm2024_conference}

\appendix
\newpage
\section{Data}
\label{sec:data}

\begin{figure}[H]
    \centering
    \begin{tabular}{cc}
     \includegraphics[width=5cm]{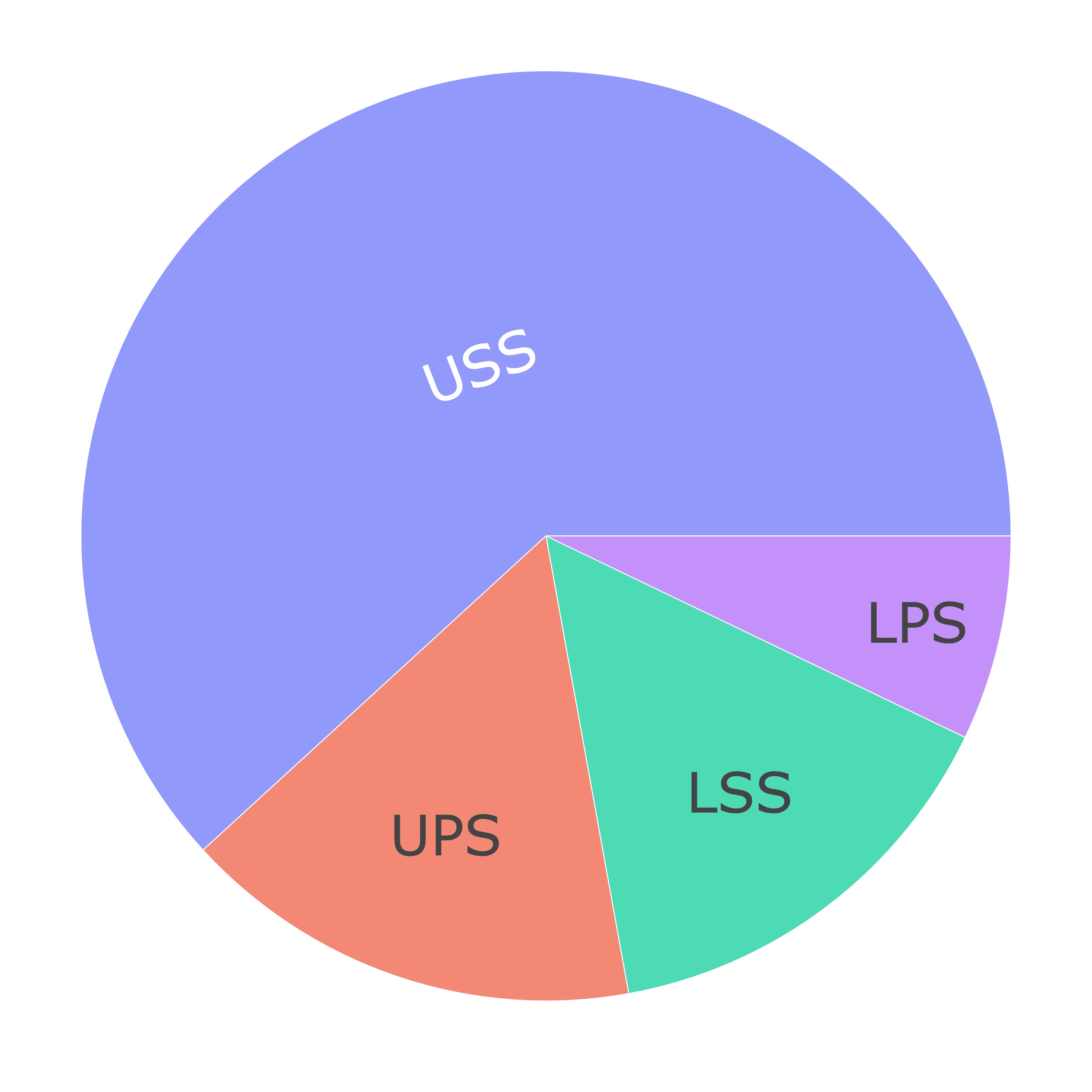} & \includegraphics[width=5cm]{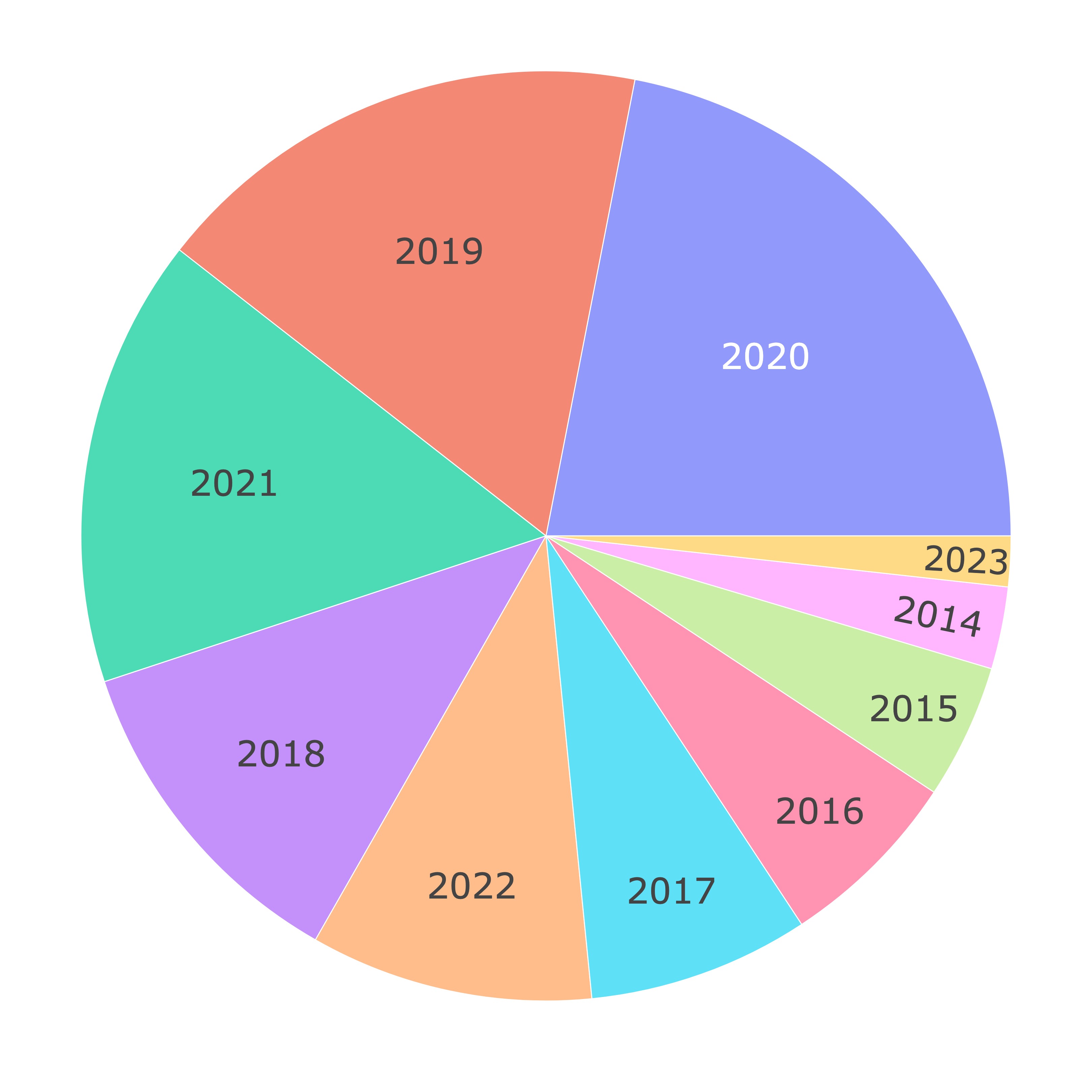}
     \\
     (a) & (b)
    \end{tabular}
    \caption{Distribution of questions across publication year and educational stage}
    \label{fig:question_distribution}
\end{figure}

\begin{figure}[h]
\begin{center}
\includegraphics[width=1\linewidth]{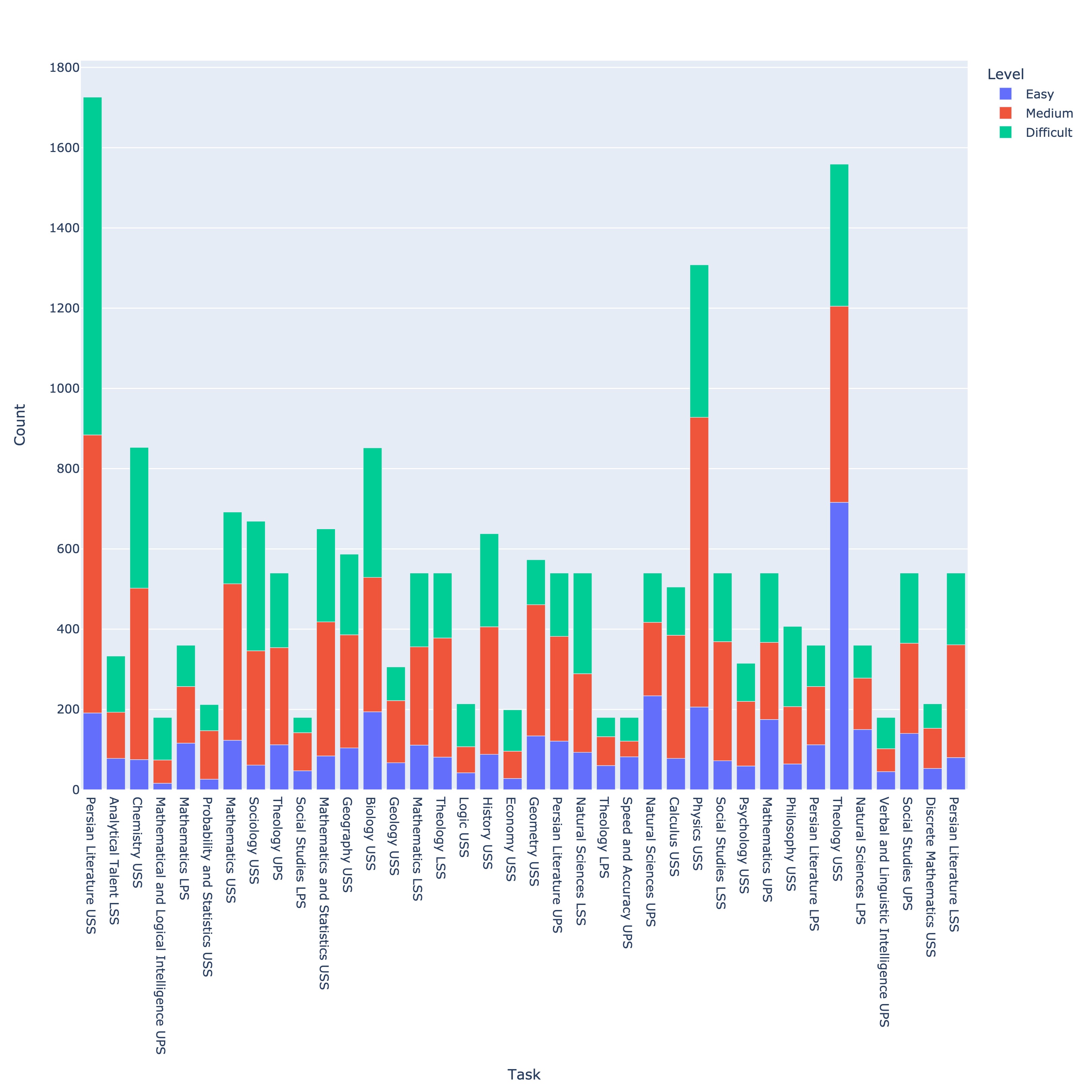}
\end{center}
\caption{Question distribution across all tasks by difficulty level}
\label{fig:question_dist_all_by_difficulty}
\end{figure}


\begin{table}[H]
\begin{center}
\scalebox{1}{
    \begin{tabular}{|c||ccc||c|}
        \hline
        \textbf{Main category} & \textbf{Easy} & \textbf{Medium} & \textbf{Difficult} & \textbf{Sum} \\ \hline \hline
        Social Science & 1480 & 2513 & 1856 & 5849 \\ \hline
        Humanities & 743 & 1963 & 1899 & 4605 \\ \hline
        Natural Science & 1019 & 2146 & 1594 & 4759 \\ \hline
        Mathematics & 998 & 2254 & 1394 & 4646 \\ \hline 
        Other & 78 & 115 & 140 & 333 \\ \hline \hline
        Sum & 4318 & 8991 & 6883 & 20192 \\ \hline
    \end{tabular}}
\end{center}
\caption{Question distribution across main categories by difficulty level}
\label{table:dist_questions_main_categories_by_diff_lvl}
\end{table}

\newpage
\section{Models}
\label{sec:models}

\subsection*{GPT-4}

OpenAI has introduced GPT-4 as a large multimodal model. However, the internal structure and operational specifics of GPT-4 are proprietary and not openly disclosed.

\subsection*{GPT-3.5}

GPT-3.5, created by OpenAI, represents a significant advancement in natural language processing technology. It boasts enhanced capabilities in understanding and generating human-like text, performing impressively across various language tasks like text completion, translation, and question answering, thanks to its expansive architecture and vast training data.

\subsection*{Claude3-haiku}

The Claude 3 Haiku model, developed recently by Anthropic, are proprietary models with undisclosed architecture and training specifics.

\subsection*{Aya}

Aya, an open-source model by Cohere, is a massively multilingual generative language model capable of understanding instructions in 101 languages. Over half of these languages are classified as lower-resourced.

\subsection*{XVERSE-13B}

XVERSE-13B, created by Shenzhen Yuanxiang Technology, stands as a versatile multilingual large language model. Employing a prominent Decoder-only Transformer network structure, XVERSE-13B has a context length of 8k. It has been meticulously trained on a vast and varied dataset containing 1.4 trillion tokens, covering more than 40 languages including Chinese, English, Russian, and Spanish.

\subsection*{XVERSE-7B}

XVERSE-7B, much like its counterpart XVERSE-13B, is a multilingual large language model developed by Shenzhen Yuanxiang Technology. It shares the same mainstream Decoder-only Transformer network structure but with a slightly reduced parameter size of 7 billion.

\subsection*{PersianMind}

PersianMind is an openly available bilingual large language model. It enriches LLaMa2's vocabulary by integrating 10,000 Persian tokens and undergoes training on a dataset containing nearly 2 billion Persian tokens.

\subsection*{mT0XL}

mT0XL originates from the fine-tuning of BLOOM models using xP3, a dataset that encompasses multilingual content coupled with English prompts. The BLOOM models, which are extensive decoder-only language models, have been pre-trained on roughly 350 billion tokens. They share structural similarities with GPT-3.

\subsection*{mGPT}

mGPT is an open-source adaptation of GPT-3, having multilingual capabilities. It undergoes pretraining across 61 languages, drawn from 25 diverse language families, leveraging Wikipedia and the C4 Corpus. Additionally, mGPT offers support for Persian language integration.

\newpage
\section{Prompt template}
\label{sec:prompt_template}
\begin{figure*}[!htb]
\begin{center}
    \sBox{
\begin{otherlanguage}{farsi}
    {\textRL{
    در زیر سوالات چند گزینه‌ای (با پاسخ) در مورد
    } \color{Black}}
    {\textRL{\textbf{هندسه دوره دوم متوسطه} } \color{Blue} }
    \textRL{
     را مشاهده می‌کنید.
    }
    \vskip 0.2cm
    {
    \textRL{سوال:}
    }
    \vskip 0.3cm
    {
    \color{Blue}
    \textRL{ \textbf{ به ازای هر $m$، معادله $(m-2)x+(m+1)y=6$، معادله قطری از دایره $C$ است. اگر نقطه}}
    \color{Blue}
    }
    {
    \textRL{\textbf{$A(-1,1)$ روی دایره $C$ باشد، محیط دایره $C$ کدام است؟}}
    \color{Blue}
    }
    \vskip 0.3cm
    {
    \textRL{گزینه‌ها:}
    \color{Black}
    }
    
    \begin{itemize}
        \item[1)] {$2\sqrt{2}\pi $ \color{Blue}}
        \item[2)] {$2\pi $ \color{Blue}}
        \item[3)] {$3\pi $ \color{Blue}}
        \item[4)] {$2\sqrt{3}\pi $ \color{Blue}}
    \end{itemize}
    {
    \textRL{جواب:}
    \color{Black}
    }
\end{otherlanguage}
\vskip 1cm
 Below you will find multiple-choice questions (with answers) regarding \textcolor{Blue}{\textbf{geometry for the second grade of high school}}.
\vskip 0.2cm
Question: 
\textcolor{Blue}{\textbf{For each $m$ the equation $(m-2)x+(m+1)y=6$, is the equation of chord of a circle $C$. If the point $A(-1,1)$ lies on the circle $C$, what is the circumference of circle $C$?}}
\vskip 0.2cm
Choices:

\begin{itemize}
    \item[1)] {\color{Blue} $2\sqrt{2}\pi $ }
    \item[2)] {\color{Blue} $2\pi $ }
    \item[3)] {\color{Blue} $3\pi $ }
    \item[4)] {\color{Blue} $2\sqrt{3}\pi $ }
\end{itemize}

Answer:
\vskip 0.5cm
}
\end{center}
    
\caption{Sample prompt-0 with English translation for enhanced readability}
\label{fig:prompt0}
\end{figure*}
\newpage

\begin{figure*}[!htb]
\begin{center}
    \sBox{
\begin{otherlanguage}{farsi}
    {\textRL{
    در زیر سوالات چند گزینه‌ای (با پاسخ) در مورد
    } \color{Black}}
    {\textRL{\textbf{هندسه دوره دوم متوسطه} } \color{Blue} }
    \textRL{
     را مشاهده می‌کنید.
    }
    \vskip 0.2cm
    {
    \textRL{سوال:}
    }
    \vskip 0.3cm
    {
    \color{Blue}
    \textRL{ \textbf{ به ازای هر $m$، معادله $(m-2)x+(m+1)y=6$، معادله قطری از دایره $C$ است. اگر نقطه}}
    \color{Blue}
    }
    {
    \textRL{\textbf{$A(-1,1)$ روی دایره $C$ باشد، محیط دایره $C$ کدام است؟}}
    \color{Blue}
    }
    \vskip 0.3cm
    {
    \textRL{گزینه‌ها:}
    \color{Black}
    }
\end{otherlanguage}
    \begin{itemize}
        \item[a)] {$2\sqrt{2}\pi $ \color{Blue}}
        \item[b)] {$2\pi $ \color{Blue}}
        \item[c)] {$3\pi $ \color{Blue}}
        \item[d)] {$2\sqrt{3}\pi $ \color{Blue}}
    \end{itemize}
\begin{otherlanguage}{farsi}
    {
    \textRL{جواب:}
    \color{Black}
    }
\end{otherlanguage}
\vskip 1cm
 Below you will find multiple-choice questions (with answers) regarding \textcolor{Blue}{\textbf{geometry for the second grade of high school}}.
\vskip 0.2cm
Question: 
\textcolor{Blue}{\textbf{For each $m$ the equation $(m-2)x+(m+1)y=6$, is the equation of chord of a circle $C$. If the point $A(-1,1)$ lies on the circle $C$, what is the circumference of circle $C$?}}
\vskip 0.2cm
Choices:

\begin{itemize}
    \item[a)] {\color{Blue} $2\sqrt{2}\pi $ }
    \item[b)] {\color{Blue} $2\pi $ }
    \item[c)] {\color{Blue} $3\pi $ }
    \item[d)] {\color{Blue} $2\sqrt{3}\pi $ }
\end{itemize}

Answer:
\vskip 0.5cm
}
\end{center}
    
\caption{Sample prompt-1 with English translation for enhanced readability}
\label{fig:prompt1}
\end{figure*}
\newpage

\begin{figure*}[!htb]
\begin{center}
    \sBox{
    The following are multiple choice questions (with answer) about \textcolor{Blue}{Upper Secondary School Geometry}
    \vskip 0.2cm
\begin{otherlanguage}{farsi}
    {
    \textRL{سوال:}
    }
    \vskip 0.3cm
    {
    \color{Blue}
    \textRL{ \textbf{ به ازای هر $m$، معادله $(m-2)x+(m+1)y=6$، معادله قطری از دایره $C$ است. اگر نقطه}}
    \color{Blue}
    }
    {
    \textRL{\textbf{$A(-1,1)$ روی دایره $C$ باشد، محیط دایره $C$ کدام است؟}}
    \color{Blue}
    }
    \vskip 0.3cm
    {
    \textRL{گزینه‌ها:}
    \color{Black}
    }
\end{otherlanguage}
    \begin{itemize}
        \item[a)] {$2\sqrt{2}\pi $ \color{Blue}}
        \item[b)] {$2\pi $ \color{Blue}}
        \item[c)] {$3\pi $ \color{Blue}}
        \item[d)] {$2\sqrt{3}\pi $ \color{Blue}}
    \end{itemize}
\begin{otherlanguage}{farsi}
    {
    \textRL{جواب:}
    \color{Black}
    }
\end{otherlanguage}
\vskip 1cm
 Below you will find multiple-choice questions (with answers) regarding \textcolor{Blue}{\textbf{geometry for the second grade of high school}}.
\vskip 0.2cm
Question: 
\textcolor{Blue}{\textbf{For each $m$ the equation $(m-2)x+(m+1)y=6$, is the equation of chord of a circle $C$. If the point $A(-1,1)$ lies on the circle $C$, what is the circumference of circle $C$?}}
\vskip 0.2cm
Choices:

\begin{itemize}
    \item[a)] {\color{Blue} $2\sqrt{2}\pi $ }
    \item[b)] {\color{Blue} $2\pi $ }
    \item[c)] {\color{Blue} $3\pi $ }
    \item[d)] {\color{Blue} $2\sqrt{3}\pi $ }
\end{itemize}

Answer:
\vskip 0.5cm
}
\end{center}
    
\caption{Sample prompt-2 with English translation for enhanced readability}
\label{fig:prompt2}
\end{figure*}
\newpage

\begin{figure*}[!htb]
\begin{center}
    \sBox{
\begin{otherlanguage}{farsi}
    {\textRL{
    در زیر سوالات چند گزینه‌ای (با پاسخ) در مورد
    } \color{Black}}
    {\textRL{\textbf{هندسه دوره دوم متوسطه} } \color{Blue} }
    \textRL{
     را مشاهده می‌کنید.
    }
    \vskip 0.2cm
    {
    \textRL{سوال:}
    }
    \vskip 0.3cm
    {
    \color{Blue}
    \textRL{ \textbf{ به ازای هر $m$، معادله $(m-2)x+(m+1)y=6$، معادله قطری از دایره $C$ است. اگر نقطه}}
    \color{Blue}
    }
    {
    \textRL{\textbf{$A(-1,1)$ روی دایره $C$ باشد، محیط دایره $C$ کدام است؟}}
    \color{Blue}
    }
    \vskip 0.3cm
    {
    \textRL{گزینه‌ها:}
    \color{Black}
    }
\end{otherlanguage}
    \begin{itemize}
        \item[a)] {$2\sqrt{2}\pi $ \color{Blue}}
        \item[b)] {$2\pi $ \color{Blue}}
        \item[c)] {$3\pi $ \color{Blue}}
        \item[d)] {$2\sqrt{3}\pi $ \color{Blue}}
    \end{itemize}
\begin{otherlanguage}{farsi}
    {
    \textRL{جواب: بیایید قدم به قدم فکر کنیم}
    \color{Black}
    }
\end{otherlanguage}
\vskip 1cm
 Below you will find multiple-choice questions (with answers) regarding \textcolor{Blue}{\textbf{geometry for the second grade of high school}}.
\vskip 0.2cm
Question: 
\textcolor{Blue}{\textbf{For each $m$ the equation $(m-2)x+(m+1)y=6$, is the equation of chord of a circle $C$. If the point $A(-1,1)$ lies on the circle $C$, what is the circumference of circle $C$?}}
\vskip 0.2cm
Choices:

\begin{itemize}
    \item[a)] {\color{Blue} $2\sqrt{2}\pi $ }
    \item[b)] {\color{Blue} $2\pi $ }
    \item[c)] {\color{Blue} $3\pi $ }
    \item[d)] {\color{Blue} $2\sqrt{3}\pi $ }
\end{itemize}

Answer: Let's think step by step
\vskip 0.5cm
}
\end{center}
    
\caption{Sample CoT prompt with English translation for enhanced readability}
\label{fig:prompt-cot}
\end{figure*}
\newpage

\section{Prompt \& CoT}

\begin{table}[H]
\begin{center}
\scalebox{1}{
    \begin{tabular}{|c|c|c|c|}
        \hline
        \textbf{Main Categories} & \textbf{Prompt-0} & \textbf{Prompt-1} & \textbf{Prompt-2} \\ \hline \hline
        Humanities & 0.32 & 0.35 & 0.36 \\ \hline
        Mathematics & 0.28 & 0.3 & 0.34 \\ \hline
        Natural Science & 0.35 & 0.36 & 0.37 \\ \hline
        Social Science & 0.43 & 0.42 & 0.42 \\ \hline
        Other & 0.45 & 0.32 & 0.27 \\ \hline \hline
        Avg on all tasks & 0.37 & 0.35 & 0.35 \\ \hline
        Avg on all questions & 0.35 & 0.36 & 0.37 \\ \hline
    \end{tabular}}
\end{center}
\caption{Accuracy of GPT-3.5 across main categories for different prompts}
\label{table:cot_0}
\end{table}

\begin{table}[H]
\begin{center}
\scalebox{1}{
    \begin{tabular}{|c|c|c|}
        \hline
        \textbf{Main Categories} & \textbf{Prompt-0} & \textbf{CoT} \\ \hline \hline
        Humanities & 0.36 & 0.28 \\ \hline
        Mathematics & 0.31 & 0.41 \\ \hline
        Other & 0.33 & 0.34 \\ \hline \hline
        Avg on all tasks & 0.33 & 0.34 \\ \hline
        Avg on all questions & 0.31 & 0.4 \\ \hline
    \end{tabular}}
\end{center}
\caption{Accuracy of GPT-3.5 across main categories for main prompt and CoT}
\label{table:cot}
\end{table}

\newpage

\section{Accuracy \& candidate choice distribution}
\label{sec:accuracy_candidate_options}

\begin{table}[H]

\label{table:labels_distribution}
\begin{center}
\scalebox{0.58}{
    \begin{tabular}{|c||c|ccccccccc||c|}

        \hline
        \textbf{Main categories} & \textbf{Human} & \textbf{GPT-4} & \textbf{Haiku} & \textbf{GPT-3.5} & \textbf{Aya} & \textbf{XVERSE-13B} & \textbf{XVERSE-7B} & \textbf{PersianMind} & \textbf{mT0XL} & \textbf{mGPT} & \textbf{Random} \\  \hline
        Humanities & 0.71 & 0.48 & 0.43 & 0.34 & 0.27 & 0.3 & 0.27 & 0.26 & 0.26 & 0.24 & 0.25 \\ \hline
        Mathematics & 0.75 & 0.4 & 0.33 & 0.32 & 0.24 & 0.25 & 0.25 & 0.24 & 0.23 & 0.23 & 0.25 \\ \hline
        Natural Science & 0.79 & 0.48 & 0.4 & 0.36 & 0.29 & 0.31 & 0.28 & 0.25 & 0.26 & 0.24 & 0.25 \\ \hline
        Social Science & 0.84 & 0.63 & 0.55 & 0.42 & 0.33 & 0.34 & 0.32 & 0.27 & 0.29 & 0.25 & 0.26 \\ \hline
        Other & 0.59 & 0.53 & 0.39 & 0.36 & 0.3 & 0.3 & 0.32 & 0.29 & 0.28 & 0.24 & 0.26 \\ \hline \hline
        Avg on all tasks & 0.74 & 0.5 & 0.42 & 0.36 & 0.28 & 0.3 & 0.29 & 0.26 & 0.26 & 0.24 & 0.25 \\ \hline
        Avg on all questions & 0.77 & 0.5 & 0.43 & 0.36 & 0.28 & 0.3 & 0.28 & 0.26 & 0.26 & 0.24 & 0.25 \\ \hline

    \end{tabular}}
\end{center}
\caption{Human vs model accuracy for main categories: model answers extracted via full answer probability} 
\label{table:acc_main_catgories_FullProbability_approach}
\end{table}

    


\begin{table}[H]
\begin{center}
\scalebox{0.58}{
    \begin{tabular}{|c||c|ccccccccc||c|}
    
        \hline
        \textbf{Main categories} & \textbf{Human} & \textbf{GPT-4} & \textbf{Haiku} & \textbf{GPT-3.5} & \textbf{Aya} & \textbf{XVERSE-13B} & \textbf{XVERSE-7B} & \textbf{PersianMind} & \textbf{mT0XL} & \textbf{mGPT} & \textbf{Random} \\  \hline 
        Humanities & 0.71 & 0.48 & 0.43 & 0.34 & 0.32 & 0.29 & 0.27 & 0.26 & 0.27 & 0.23 & 0.25 \\ \hline
        Mathematics & 0.75 & 0.4 & 0.33 & 0.32 & 0.27 & 0.26 & 0.25 & 0.25 & 0.26 & 0.23 & 0.25 \\ \hline
        Natural Science & 0.79 & 0.48 & 0.4 & 0.36 & 0.33 & 0.3 & 0.28 & 0.26 & 0.28 & 0.21 & 0.25 \\ \hline
        Social Science & 0.84 & 0.63 & 0.55 & 0.42 & 0.41 & 0.34 & 0.32 & 0.29 & 0.31 & 0.22 & 0.26 \\ \hline
        Other & 0.59 & 0.53 & 0.39 & 0.36 & 0.27 & 0.3 & 0.32 & 0.27 & 0.3 & 0.23 & 0.26 \\ \hline \hline
        Avg on all tasks & 0.74 & 0.5 & 0.42 & 0.36 & 0.32 & 0.3 & 0.29 & 0.27 & 0.28 & 0.22 & 0.25 \\ \hline
        Avg on all questions & 0.77 & 0.5 & 0.43 & 0.36 & 0.34 & 0.3 & 0.29 & 0.27 & 0.28 & 0.22 & 0.25 \\ \hline

    \end{tabular}}
\end{center}
\caption{Human vs model accuracy for main categories: model answers extracted via single token probability} 
\label{table:acc_main_catgories_probability_approach}
\end{table}

\newpage
\begin{table}[!htb]
\begin{center}
\scalebox{0.5}{
    \begin{tabular}{|c|c|ccccccccc|c|}
    \hline
        \textbf{Task} & \textbf{Human} & \textbf{GPT-4} & \textbf{Haiku} & \textbf{GPT-3.5} & \textbf{Aya} & \textbf{XVERSE-13B} & \textbf{XVERSE-7B} & \textbf{PersianMind} & \textbf{mT0XL} & \textbf{mGPT} & \textbf{Random} \\ \hline\hline 
        Analytical Talent LSS & 0.59 & 0.53 & 0.39 & 0.36 & 0.29 & 0.29 & 0.32 & 0.27 & 0.28 & 0.28 & 0.26 \\ \hline
        Calculus USS & 0.85 & 0.34 & 0.29 & 0.26 & 0.26 & 0.25 & 0.26 & 0.21 & 0.25 & 0.27 & 0.29 \\ \hline
        Chemistry USS & 0.73 & 0.36 & 0.3 & 0.31 & 0.3 & 0.28 & 0.27 & 0.26 & 0.28 & 0.26 & 0.24 \\ \hline
        Discrete Mathematics USS & 0.78 & 0.29 & 0.27 & 0.22 & 0.24 & 0.24 & 0.23 & 0.29 & 0.26 & 0.27 & 0.3 \\ \hline
        Economy USS & 0.69 & 0.62 & 0.51 & 0.4 & 0.47 & 0.33 & 0.25 & 0.28 & 0.31 & 0.25 & 0.29 \\ \hline
        Geography USS & 0.85 & 0.56 & 0.5 & 0.38 & 0.36 & 0.3 & 0.31 & 0.3 & 0.28 & 0.26 & 0.25 \\ \hline
        Geology USS & 0.89 & 0.56 & 0.52 & 0.39 & 0.36 & 0.32 & 0.23 & 0.27 & 0.25 & 0.25 & 0.26 \\ \hline
        Geometry USS & 0.83 & 0.34 & 0.32 & 0.3 & 0.28 & 0.27 & 0.28 & 0.28 & 0.25 & 0.23 & 0.24 \\ \hline
        History USS & 0.85 & 0.55 & 0.47 & 0.37 & 0.33 & 0.33 & 0.29 & 0.31 & 0.27 & 0.25 & 0.24 \\ \hline
        Logic USS & 0.58 & 0.43 & 0.4 & 0.31 & 0.29 & 0.33 & 0.29 & 0.31 & 0.22 & 0.21 & 0.26 \\ \hline
        Mathematical and Logical Intelligence UPS & 0.44 & 0.45 & 0.28 & 0.26 & 0.29 & 0.22 & 0.22 & 0.24 & 0.21 & 0.28 & 0.18 \\ \hline
        Mathematics LPS & 0.77 & 0.56 & 0.38 & 0.39 & 0.31 & 0.33 & 0.32 & 0.22 & 0.27 & 0.25 & 0.26 \\ \hline
        Mathematics LSS & 0.77 & 0.4 & 0.36 & 0.32 & 0.31 & 0.31 & 0.26 & 0.29 & 0.25 & 0.25 & 0.26 \\ \hline
        Mathematics UPS & 0.66 & 0.49 & 0.35 & 0.35 & 0.29 & 0.27 & 0.27 & 0.26 & 0.27 & 0.27 & 0.28 \\ \hline
        Mathematics USS & 0.84 & 0.34 & 0.29 & 0.31 & 0.27 & 0.24 & 0.25 & 0.27 & 0.25 & 0.25 & 0.26 \\ \hline
        Mathematics and Statistics USS & 0.63 & 0.42 & 0.36 & 0.34 & 0.27 & 0.28 & 0.28 & 0.28 & 0.27 & 0.27 & 0.23 \\ \hline
        Natural Sciences LPS & 0.85 & 0.79 & 0.71 & 0.61 & 0.5 & 0.48 & 0.4 & 0.33 & 0.38 & 0.28 & 0.25 \\ \hline
        Natural Sciences LSS & 0.72 & 0.59 & 0.46 & 0.39 & 0.33 & 0.32 & 0.29 & 0.3 & 0.28 & 0.25 & 0.23 \\ \hline
        Natural Sciences UPS & 0.82 & 0.74 & 0.63 & 0.5 & 0.43 & 0.39 & 0.32 & 0.29 & 0.3 & 0.24 & 0.26 \\ \hline
        Persian Literature LPS & 0.83 & 0.66 & 0.55 & 0.42 & 0.41 & 0.39 & 0.3 & 0.29 & 0.33 & 0.29 & 0.25 \\ \hline
        Persian Literature LSS & 0.77 & 0.51 & 0.46 & 0.34 & 0.3 & 0.29 & 0.28 & 0.26 & 0.3 & 0.24 & 0.25 \\ \hline
        Persian Literature UPS & 0.8 & 0.57 & 0.53 & 0.38 & 0.31 & 0.35 & 0.28 & 0.26 & 0.31 & 0.26 & 0.24 \\ \hline
        Persian Literature USS & 0.59 & 0.35 & 0.32 & 0.27 & 0.29 & 0.25 & 0.26 & 0.26 & 0.26 & 0.25 & 0.25 \\ \hline
        Philosophy USS & 0.61 & 0.54 & 0.51 & 0.39 & 0.36 & 0.31 & 0.26 & 0.31 & 0.33 & 0.27 & 0.24 \\ \hline
        Physics USS & 0.83 & 0.39 & 0.32 & 0.29 & 0.27 & 0.28 & 0.24 & 0.27 & 0.27 & 0.25 & 0.26 \\ \hline
        Probability and Statistics USS & 0.77 & 0.39 & 0.35 & 0.31 & 0.25 & 0.25 & 0.24 & 0.2 & 0.23 & 0.23 & 0.23 \\ \hline
        Psychology USS & 0.79 & 0.63 & 0.5 & 0.4 & 0.4 & 0.32 & 0.35 & 0.32 & 0.35 & 0.22 & 0.24 \\ \hline
        Social Studies LPS & 0.94 & 0.85 & 0.77 & 0.67 & 0.62 & 0.59 & 0.44 & 0.45 & 0.38 & 0.25 & 0.23 \\ \hline
        Social Studies LSS & 0.86 & 0.67 & 0.57 & 0.43 & 0.39 & 0.35 & 0.34 & 0.29 & 0.31 & 0.23 & 0.26 \\ \hline
        Social Studies UPS & 0.89 & 0.73 & 0.7 & 0.52 & 0.46 & 0.36 & 0.36 & 0.38 & 0.36 & 0.26 & 0.26 \\ \hline
        Sociology USS & 0.82 & 0.48 & 0.44 & 0.33 & 0.34 & 0.32 & 0.28 & 0.27 & 0.26 & 0.27 & 0.24 \\ \hline
        Speed and Accuracy UPS & 0.84 & 0.38 & 0.32 & 0.36 & 0.27 & 0.29 & 0.26 & 0.26 & 0.28 & 0.26 & 0.17 \\ \hline
        Theology LPS & 0.88 & 0.84 & 0.71 & 0.61 & 0.54 & 0.48 & 0.44 & 0.34 & 0.51 & 0.21 & 0.31 \\ \hline
        Theology LSS & 0.88 & 0.69 & 0.61 & 0.45 & 0.46 & 0.38 & 0.34 & 0.31 & 0.35 & 0.26 & 0.29 \\ \hline
        Theology UPS & 0.9 & 0.79 & 0.68 & 0.54 & 0.49 & 0.41 & 0.38 & 0.34 & 0.35 & 0.25 & 0.28 \\ \hline
        Theology USS & 0.8 & 0.53 & 0.45 & 0.34 & 0.37 & 0.28 & 0.28 & 0.26 & 0.3 & 0.24 & 0.24 \\ \hline
        Verbal and Linguistic Intelligence UPS & 0.63 & 0.55 & 0.49 & 0.43 & 0.36 & 0.37 & 0.27 & 0.28 & 0.31 & 0.33 & 0.26 \\ \hline
        Biology USS & 0.73 & 0.32 & 0.27 & 0.3 & 0.29 & 0.27 & 0.28 & 0.3 & 0.26 & 0.22 & 0.25 \\ \hline \hline
        Avg on all tasks & 0.77 & 0.53 & 0.46 & 0.38 & 0.35 & 0.32 & 0.3 & 0.29 & 0.29 & 0.25 & 0.25 \\ \hline
        Avg on all questions & 0.77 & 0.5 & 0.43 & 0.36 & 0.34 & 0.31 & 0.29 & 0.28 & 0.29 & 0.25 & 0.25 \\ \hline
    \end{tabular}
    }
\end{center}
\caption{Accuracy of humans and different models across all subjects and educational stages: model answers extracted via Regex method. LPS: Lower Primary School, UPS: Upper Primary School, LSS: Lower Secondary School, USS: Upper Secondary School} 
\label{table:Acc_regex_all_subjects}
\end{table}

\newpage
\begin{table}[!htb]
\begin{center}
\scalebox{0.5}{
    \begin{tabular}{|c|c|ccccccccc|c|}
    
        \hline
        \textbf{Task} & \textbf{Human} & \textbf{GPT-4} & \textbf{Haiku} & \textbf{GPT-3.5} & \textbf{Aya} & \textbf{XVERSE-13B} & \textbf{XVERSE-7B} & \textbf{PersianMind} & \textbf{mT0XL} & \textbf{mGPT} & \textbf{Random} \\ \hline \hline
        Analytical Talent LSS & 0.59 & 0.53 & 0.39 & 0.36 & 0.27 & 0.3 & 0.32 & 0.27 & 0.3 & 0.23 & 0.26 \\ \hline
        Calculus USS & 0.85 & 0.34 & 0.29 & 0.26 & 0.26 & 0.28 & 0.3 & 0.24 & 0.25 & 0.25 & 0.29 \\ \hline
        Chemistry USS & 0.73 & 0.36 & 0.3 & 0.31 & 0.3 & 0.29 & 0.27 & 0.24 & 0.27 & 0.24 & 0.24 \\ \hline
        Discrete Mathematics USS & 0.78 & 0.29 & 0.27 & 0.22 & 0.24 & 0.27 & 0.26 & 0.26 & 0.24 & 0.27 & 0.3 \\ \hline
        Economy USS & 0.69 & 0.62 & 0.51 & 0.4 & 0.43 & 0.33 & 0.24 & 0.3 & 0.25 & 0.22 & 0.29 \\ \hline
        Geography USS & 0.85 & 0.56 & 0.5 & 0.38 & 0.36 & 0.32 & 0.31 & 0.29 & 0.28 & 0.24 & 0.25 \\ \hline
        Geology USS & 0.89 & 0.56 & 0.52 & 0.39 & 0.36 & 0.32 & 0.22 & 0.2 & 0.25 & 0.2 & 0.26 \\ \hline
        Geometry USS & 0.83 & 0.34 & 0.32 & 0.3 & 0.28 & 0.25 & 0.25 & 0.24 & 0.26 & 0.22 & 0.24 \\ \hline
        History USS & 0.85 & 0.55 & 0.47 & 0.37 & 0.34 & 0.31 & 0.28 & 0.27 & 0.27 & 0.21 & 0.24 \\ \hline
        Logic USS & 0.58 & 0.43 & 0.4 & 0.31 & 0.29 & 0.3 & 0.29 & 0.29 & 0.28 & 0.29 & 0.26 \\ \hline
        Mathematical and Logical Intelligence UPS & 0.44 & 0.45 & 0.28 & 0.26 & 0.3 & 0.27 & 0.24 & 0.29 & 0.29 & 0.27 & 0.18 \\ \hline
        Mathematics LPS & 0.77 & 0.56 & 0.38 & 0.39 & 0.29 & 0.25 & 0.27 & 0.27 & 0.27 & 0.22 & 0.26 \\ \hline
        Mathematics LSS & 0.77 & 0.4 & 0.36 & 0.32 & 0.3 & 0.23 & 0.25 & 0.25 & 0.26 & 0.21 & 0.26 \\ \hline
        Mathematics UPS & 0.66 & 0.49 & 0.35 & 0.35 & 0.27 & 0.24 & 0.23 & 0.23 & 0.23 & 0.21 & 0.28 \\ \hline
        Mathematics USS & 0.84 & 0.34 & 0.29 & 0.31 & 0.27 & 0.23 & 0.23 & 0.24 & 0.28 & 0.23 & 0.26 \\ \hline
        Mathematics and Statistics USS & 0.63 & 0.42 & 0.36 & 0.34 & 0.26 & 0.31 & 0.27 & 0.26 & 0.28 & 0.23 & 0.23 \\ \hline
        Natural Sciences LPS & 0.85 & 0.79 & 0.71 & 0.61 & 0.51 & 0.44 & 0.39 & 0.34 & 0.33 & 0.19 & 0.25 \\ \hline
        Natural Sciences LSS & 0.72 & 0.59 & 0.46 & 0.39 & 0.34 & 0.31 & 0.29 & 0.27 & 0.29 & 0.18 & 0.23 \\ \hline
        Natural Sciences UPS & 0.82 & 0.74 & 0.63 & 0.5 & 0.42 & 0.37 & 0.32 & 0.26 & 0.27 & 0.19 & 0.26 \\ \hline
        Persian Literature LPS & 0.83 & 0.66 & 0.55 & 0.42 & 0.41 & 0.37 & 0.3 & 0.26 & 0.34 & 0.19 & 0.25 \\ \hline
        Persian Literature LSS & 0.77 & 0.51 & 0.46 & 0.34 & 0.32 & 0.3 & 0.28 & 0.28 & 0.26 & 0.26 & 0.25 \\ \hline
        Persian Literature UPS & 0.8 & 0.57 & 0.53 & 0.38 & 0.36 & 0.32 & 0.28 & 0.25 & 0.31 & 0.2 & 0.24 \\ \hline
        Persian Literature USS & 0.59 & 0.35 & 0.32 & 0.27 & 0.28 & 0.25 & 0.26 & 0.25 & 0.25 & 0.22 & 0.25 \\ \hline
        Philosophy USS & 0.61 & 0.54 & 0.51 & 0.39 & 0.34 & 0.31 & 0.26 & 0.29 & 0.26 & 0.27 & 0.24 \\ \hline
        Physics USS & 0.83 & 0.39 & 0.32 & 0.29 & 0.27 & 0.27 & 0.25 & 0.27 & 0.28 & 0.22 & 0.26 \\ \hline
        Probability and Statistics USS & 0.77 & 0.39 & 0.35 & 0.31 & 0.24 & 0.29 & 0.21 & 0.25 & 0.25 & 0.22 & 0.23 \\ \hline
        Psychology USS & 0.79 & 0.63 & 0.5 & 0.4 & 0.37 & 0.35 & 0.36 & 0.3 & 0.33 & 0.24 & 0.24 \\ \hline
        Social Studies LPS & 0.94 & 0.85 & 0.77 & 0.67 & 0.59 & 0.53 & 0.43 & 0.36 & 0.32 & 0.17 & 0.23 \\ \hline
        Social Studies LSS & 0.86 & 0.67 & 0.57 & 0.43 & 0.39 & 0.35 & 0.34 & 0.29 & 0.31 & 0.22 & 0.26 \\ \hline
        Social Studies UPS & 0.89 & 0.73 & 0.7 & 0.52 & 0.48 & 0.36 & 0.37 & 0.34 & 0.33 & 0.2 & 0.26 \\ \hline
        Sociology USS & 0.82 & 0.48 & 0.44 & 0.33 & 0.34 & 0.29 & 0.29 & 0.26 & 0.28 & 0.22 & 0.24 \\ \hline
        Speed and Accuracy UPS & 0.84 & 0.38 & 0.32 & 0.36 & 0.24 & 0.23 & 0.26 & 0.25 & 0.26 & 0.21 & 0.17 \\ \hline
        Theology LPS & 0.88 & 0.84 & 0.71 & 0.61 & 0.56 & 0.44 & 0.44 & 0.33 & 0.46 & 0.13 & 0.31 \\ \hline
        Theology LSS & 0.88 & 0.69 & 0.61 & 0.45 & 0.48 & 0.38 & 0.32 & 0.31 & 0.34 & 0.23 & 0.29 \\ \hline
        Theology UPS & 0.9 & 0.79 & 0.68 & 0.54 & 0.49 & 0.4 & 0.38 & 0.33 & 0.39 & 0.22 & 0.28 \\ \hline
        Theology USS & 0.8 & 0.53 & 0.45 & 0.34 & 0.37 & 0.28 & 0.28 & 0.26 & 0.26 & 0.23 & 0.24 \\ \hline
        Verbal and Linguistic Intelligence UPS & 0.63 & 0.55 & 0.49 & 0.43 & 0.43 & 0.35 & 0.27 & 0.27 & 0.34 & 0.22 & 0.26 \\ \hline
        Biology USS & 0.73 & 0.32 & 0.27 & 0.3 & 0.28 & 0.27 & 0.29 & 0.25 & 0.25 & 0.23 & 0.25 \\ \hline \hline
        Avg on all tasks & 0.77 & 0.53 & 0.46 & 0.38 & 0.35 & 0.31 & 0.29 & 0.27 & 0.29 & 0.22 & 0.25 \\ \hline
        Avg on all questions & 0.77 & 0.5 & 0.43 & 0.36 & 0.34 & 0.3 & 0.29 & 0.27 & 0.28 & 0.22 & 0.25 \\ \hline

    \end{tabular}
    }
\end{center}
\caption{Accuracy of human and different models across all subjects and educational stages: model answers extracted via single token probability. LPS: Lower Primary School, UPS: Upper Primary School, LSS: Lower Secondary School, USS: Upper Secondary School}
\label{table:Acc_prob_all_subjects}
\end{table}

\newpage

\newpage
\begin{table}[!htb]
\begin{center}
\scalebox{0.5}{
    \begin{tabular}{|c|c|ccccccccc|c|}
    \hline
        \textbf{Task} & \textbf{Human} & \textbf{GPT-4} & \textbf{Haiku} & \textbf{GPT-3.5} & \textbf{Aya} & \textbf{XVERSE-13B} & \textbf{XVERSE-7B} & \textbf{PersianMind} & \textbf{mT0XL} & \textbf{mGPT} & \textbf{Random} \\ \hline \hline
        Analytical Talent LSS & 0.59 & 0.53 & 0.39 & 0.36 & 0.3 & 0.3 & 0.32 & 0.29 & 0.28 & 0.24 & 0.26 \\ \hline
        Calculus USS & 0.85 & 0.34 & 0.29 & 0.26 & 0.21 & 0.25 & 0.28 & 0.22 & 0.2 & 0.21 & 0.29 \\ \hline
        Chemistry USS & 0.73 & 0.36 & 0.3 & 0.31 & 0.28 & 0.29 & 0.28 & 0.26 & 0.26 & 0.24 & 0.24 \\ \hline
        Discrete Mathematics USS & 0.78 & 0.29 & 0.27 & 0.22 & 0.22 & 0.27 & 0.26 & 0.29 & 0.28 & 0.26 & 0.3 \\ \hline
        Economy USS & 0.69 & 0.62 & 0.51 & 0.4 & 0.27 & 0.33 & 0.23 & 0.24 & 0.23 & 0.23 & 0.29 \\ \hline
        Geography USS & 0.85 & 0.56 & 0.5 & 0.38 & 0.31 & 0.32 & 0.31 & 0.29 & 0.26 & 0.26 & 0.25 \\ \hline
        Geology USS & 0.89 & 0.56 & 0.52 & 0.39 & 0.3 & 0.32 & 0.22 & 0.27 & 0.26 & 0.23 & 0.26 \\ \hline
        Geometry USS & 0.83 & 0.34 & 0.32 & 0.3 & 0.25 & 0.25 & 0.24 & 0.23 & 0.22 & 0.23 & 0.24 \\ \hline
        History USS & 0.85 & 0.55 & 0.47 & 0.37 & 0.28 & 0.31 & 0.29 & 0.26 & 0.26 & 0.24 & 0.24 \\ \hline
        Logic USS & 0.58 & 0.43 & 0.4 & 0.31 & 0.27 & 0.32 & 0.29 & 0.22 & 0.21 & 0.25 & 0.26 \\ \hline
        Mathematical and Logical Intelligence UPS & 0.44 & 0.45 & 0.28 & 0.26 & 0.25 & 0.26 & 0.23 & 0.26 & 0.26 & 0.26 & 0.18 \\ \hline
        Mathematics LPS & 0.77 & 0.56 & 0.38 & 0.39 & 0.23 & 0.26 & 0.28 & 0.25 & 0.25 & 0.25 & 0.26 \\ \hline
        Mathematics LSS & 0.77 & 0.4 & 0.36 & 0.32 & 0.27 & 0.24 & 0.24 & 0.24 & 0.24 & 0.24 & 0.26 \\ \hline
        Mathematics UPS & 0.66 & 0.49 & 0.35 & 0.35 & 0.26 & 0.23 & 0.23 & 0.27 & 0.22 & 0.22 & 0.28 \\ \hline
        Mathematics USS & 0.84 & 0.34 & 0.29 & 0.31 & 0.22 & 0.22 & 0.22 & 0.22 & 0.22 & 0.24 & 0.26 \\ \hline
        Mathematics and Statistics USS & 0.63 & 0.42 & 0.36 & 0.34 & 0.23 & 0.31 & 0.27 & 0.25 & 0.21 & 0.23 & 0.23 \\ \hline
        Natural Sciences LPS & 0.85 & 0.79 & 0.71 & 0.61 & 0.39 & 0.45 & 0.4 & 0.27 & 0.28 & 0.21 & 0.25 \\ \hline
        Natural Sciences LSS & 0.72 & 0.59 & 0.46 & 0.39 & 0.3 & 0.31 & 0.29 & 0.25 & 0.29 & 0.25 & 0.23 \\ \hline
        Natural Sciences UPS & 0.82 & 0.74 & 0.63 & 0.5 & 0.31 & 0.39 & 0.33 & 0.26 & 0.23 & 0.22 & 0.26 \\ \hline
        Persian Literature LPS & 0.83 & 0.66 & 0.55 & 0.42 & 0.32 & 0.37 & 0.3 & 0.25 & 0.25 & 0.22 & 0.25 \\ \hline
        Persian Literature LSS & 0.77 & 0.51 & 0.46 & 0.34 & 0.26 & 0.3 & 0.28 & 0.23 & 0.26 & 0.23 & 0.25 \\ \hline
        Persian Literature UPS & 0.8 & 0.57 & 0.53 & 0.38 & 0.27 & 0.33 & 0.28 & 0.26 & 0.28 & 0.25 & 0.24 \\ \hline
        Persian Literature USS & 0.59 & 0.35 & 0.32 & 0.27 & 0.24 & 0.25 & 0.26 & 0.26 & 0.25 & 0.22 & 0.25 \\ \hline
        Philosophy USS & 0.61 & 0.54 & 0.51 & 0.39 & 0.27 & 0.32 & 0.26 & 0.27 & 0.27 & 0.3 & 0.24 \\ \hline
        Physics USS & 0.83 & 0.39 & 0.32 & 0.29 & 0.25 & 0.26 & 0.24 & 0.23 & 0.27 & 0.23 & 0.26 \\ \hline
        Probability and Statistics USS & 0.77 & 0.39 & 0.35 & 0.31 & 0.22 & 0.28 & 0.21 & 0.23 & 0.27 & 0.2 & 0.23 \\ \hline
        Psychology USS & 0.79 & 0.63 & 0.5 & 0.4 & 0.3 & 0.34 & 0.36 & 0.3 & 0.23 & 0.23 & 0.24 \\ \hline
        Social Studies LPS & 0.94 & 0.85 & 0.77 & 0.67 & 0.47 & 0.53 & 0.44 & 0.32 & 0.29 & 0.25 & 0.23 \\ \hline
        Social Studies LSS & 0.86 & 0.67 & 0.57 & 0.43 & 0.35 & 0.35 & 0.34 & 0.28 & 0.3 & 0.22 & 0.26 \\ \hline
        Social Studies UPS & 0.89 & 0.73 & 0.7 & 0.52 & 0.37 & 0.37 & 0.37 & 0.28 & 0.31 & 0.26 & 0.26 \\ \hline
        Sociology USS & 0.82 & 0.48 & 0.44 & 0.33 & 0.25 & 0.27 & 0.28 & 0.26 & 0.26 & 0.25 & 0.24 \\ \hline
        Speed and Accuracy UPS & 0.84 & 0.38 & 0.32 & 0.36 & 0.25 & 0.25 & 0.28 & 0.22 & 0.23 & 0.24 & 0.17 \\ \hline
        Theology LPS & 0.88 & 0.84 & 0.71 & 0.61 & 0.38 & 0.46 & 0.44 & 0.23 & 0.38 & 0.24 & 0.31 \\ \hline
        Theology LSS & 0.88 & 0.69 & 0.61 & 0.45 & 0.36 & 0.37 & 0.33 & 0.25 & 0.34 & 0.27 & 0.29 \\ \hline
        Theology UPS & 0.9 & 0.79 & 0.68 & 0.54 & 0.37 & 0.4 & 0.38 & 0.27 & 0.29 & 0.24 & 0.28 \\ \hline
        Theology USS & 0.8 & 0.53 & 0.45 & 0.34 & 0.3 & 0.28 & 0.28 & 0.24 & 0.29 & 0.26 & 0.24 \\ \hline
        Verbal and Linguistic Intelligence UPS & 0.63 & 0.55 & 0.49 & 0.43 & 0.35 & 0.38 & 0.26 & 0.29 & 0.31 & 0.27 & 0.26 \\ \hline
        Biology USS & 0.73 & 0.32 & 0.27 & 0.3 & 0.28 & 0.28 & 0.28 & 0.25 & 0.25 & 0.29 & 0.25 \\ \hline \hline
        Avg on all tasks & 0.77 & 0.53 & 0.46 & 0.38 & 0.29 & 0.32 & 0.29 & 0.26 & 0.26 & 0.24 & 0.25 \\ \hline
        Avg on all questions & 0.77 & 0.5 & 0.43 & 0.36 & 0.28 & 0.3 & 0.28 & 0.26 & 0.26 & 0.24 & 0.25 \\ \hline
        

    \end{tabular}
    }
\end{center}
\caption{Accuracy of human and different models across all subjects and educational stages: model answers extracted via full answer probability. LPS: Lower Primary School, UPS: Upper Primary School, LSS: Lower Secondary School, USS: Upper Secondary School}
\label{table:Acc_fullprob_all_subjects}
\end{table}
\newpage
\begin{figure}[H]
\begin{center}
\includegraphics[width=1\linewidth]{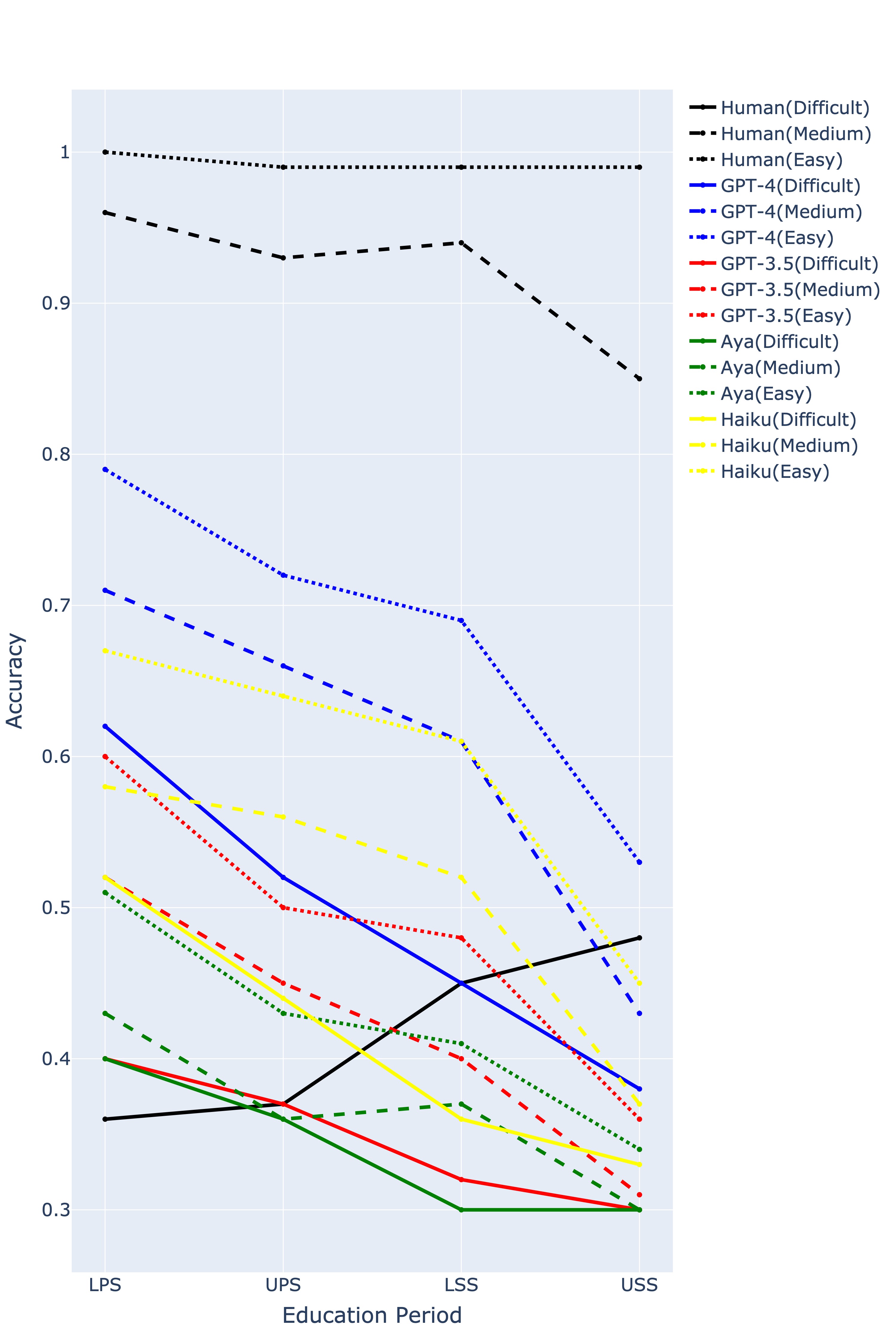}

\end{center}
\caption{Accuracy of models and humans across three difficulty levels, segmented by question educational stage.}
\label{fig:question_dist_educational_stage}
\end{figure}
\begin{figure}[H]
\begin{center}
\includegraphics[width=1\linewidth]{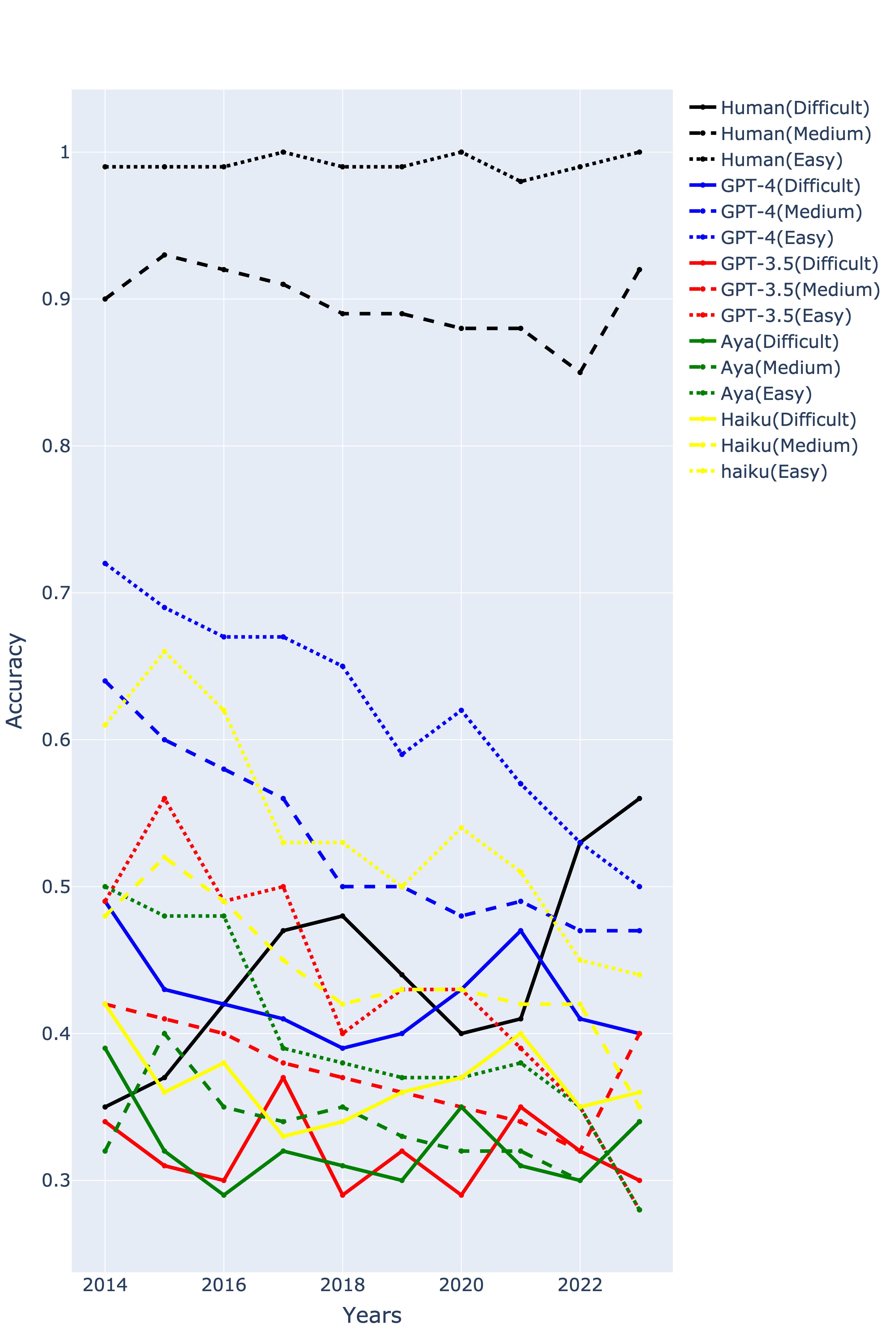}
\end{center}
\caption{Accuracy of models and humans across three difficulty levels, segmented by question publication year.}
\label{fig:question_dist_published_year}
\end{figure}
\newpage

\newpage
\begin{table}[!htb]
\begin{center}
\scalebox{0.5}{
    \begin{tabular}{|c|c|cccccccccc|c|}

        \hline
        \textbf{Main category} & \textbf{Difficulty level} & \textbf{Human} & \textbf{GPT-4} & \textbf{Haiku} & \textbf{GPT-3.5} & \textbf{Aya} & \textbf{XVERSE-13B} & \textbf{XVERSE-7B} & \textbf{PersianMind} & \textbf{mT0XL} & \textbf{mGPT} & \textbf{Random} \\ \hline \hline
        Humanities & Difficult & 0.35 & 0.39 & 0.36 & 0.31 & 0.3 & 0.28 & 0.26 & 0.27 & 0.27 & 0.26 & 0.24 \\ \hline
        Humanities & Easy & 0.99 & 0.61 & 0.54 & 0.39 & 0.36 & 0.33 & 0.3 & 0.28 & 0.3 & 0.24 & 0.26 \\ \hline
        Humanities & Medium & 0.86 & 0.5 & 0.45 & 0.34 & 0.32 & 0.32 & 0.28 & 0.28 & 0.29 & 0.26 & 0.25 \\ \hline
        Mathematics & Difficult & 0.4 & 0.36 & 0.3 & 0.3 & 0.29 & 0.26 & 0.27 & 0.25 & 0.27 & 0.26 & 0.24 \\ \hline
        Mathematics & Easy & 0.99 & 0.49 & 0.36 & 0.35 & 0.28 & 0.29 & 0.28 & 0.28 & 0.26 & 0.26 & 0.26 \\ \hline
        Mathematics & Medium & 0.84 & 0.38 & 0.33 & 0.32 & 0.27 & 0.27 & 0.25 & 0.26 & 0.25 & 0.25 & 0.26 \\ \hline
        Natural Science & Difficult & 0.5 & 0.39 & 0.32 & 0.32 & 0.31 & 0.28 & 0.27 & 0.27 & 0.27 & 0.24 & 0.26 \\ \hline
        Natural Science & Easy & 1.0 & 0.66 & 0.58 & 0.48 & 0.41 & 0.38 & 0.32 & 0.33 & 0.3 & 0.25 & 0.26 \\ \hline
        Natural Science & Medium & 0.91 & 0.45 & 0.38 & 0.33 & 0.3 & 0.31 & 0.26 & 0.28 & 0.28 & 0.25 & 0.24 \\ \hline
        Social Science & Difficult & 0.55 & 0.53 & 0.46 & 0.36 & 0.37 & 0.3 & 0.27 & 0.28 & 0.3 & 0.25 & 0.26 \\ \hline
        Social Science & Easy & 0.99 & 0.68 & 0.61 & 0.48 & 0.46 & 0.38 & 0.37 & 0.34 & 0.36 & 0.24 & 0.25 \\ \hline
        Social Science & Medium & 0.95 & 0.66 & 0.57 & 0.43 & 0.42 & 0.34 & 0.34 & 0.31 & 0.32 & 0.26 & 0.26 \\ \hline
        Other & Difficult & 0.21 & 0.44 & 0.29 & 0.28 & 0.23 & 0.23 & 0.26 & 0.23 & 0.28 & 0.34 & 0.24 \\ \hline
        Other & Easy & 0.99 & 0.67 & 0.51 & 0.46 & 0.37 & 0.36 & 0.37 & 0.35 & 0.28 & 0.17 & 0.24 \\ \hline
        Other & Medium & 0.78 & 0.55 & 0.43 & 0.4 & 0.31 & 0.32 & 0.35 & 0.26 & 0.27 & 0.28 & 0.3 \\ \hline \hline
        Avg on all tasks & Avg on all tasks & 0.75 & 0.52 & 0.43 & 0.37 & 0.33 & 0.31 & 0.3 & 0.28 & 0.29 & 0.25 & 0.25 \\ \hline
        Avg on all questions & Avg on all questions & 0.77 & 0.5 & 0.43 & 0.36 & 0.34 & 0.31 & 0.29 & 0.28 & 0.29 & 0.25 & 0.25 \\ \hline

    \end{tabular}}
\end{center}
\caption{Human vs model accuracy for all tasks: model answers extracted via regex}  
\label{table:acc_over_different_diff_lvls_reg}
\end{table}

\begin{table}[H]
\begin{center}
\scalebox{0.5}{
    \begin{tabular}{|c|c|c|ccccccccc|c|}
        \hline
        \textbf{Main category} & \textbf{Difficulty level} & \textbf{Human} & \textbf{GPT-4} & \textbf{Haiku} & \textbf{GPT-3.5} & \textbf{Aya} & \textbf{XVERSE-13B} & \textbf{XVERSE-7B} & \textbf{PersianMind} & \textbf{mT0XL} & \textbf{mGPT} & \textbf{Random} \\ \hline \hline
        Humanities & Difficult & 0.35 & 0.39 & 0.36 & 0.31 & 0.3 & 0.27 & 0.26 & 0.26 & 0.26 & 0.23 & 0.24 \\ \hline
        Humanities & Easy & 0.99 & 0.61 & 0.54 & 0.39 & 0.38 & 0.31 & 0.29 & 0.28 & 0.29 & 0.24 & 0.26 \\ \hline
        Humanities & Medium & 0.86 & 0.5 & 0.45 & 0.34 & 0.33 & 0.31 & 0.28 & 0.26 & 0.28 & 0.22 & 0.25 \\ \hline
        Mathematics & Difficult & 0.4 & 0.36 & 0.3 & 0.3 & 0.28 & 0.25 & 0.25 & 0.25 & 0.28 & 0.23 & 0.24 \\ \hline
        Mathematics & Easy & 0.99 & 0.49 & 0.36 & 0.35 & 0.27 & 0.27 & 0.26 & 0.24 & 0.26 & 0.21 & 0.26 \\ \hline
        Mathematics & Medium & 0.84 & 0.38 & 0.33 & 0.32 & 0.26 & 0.26 & 0.25 & 0.25 & 0.25 & 0.24 & 0.26 \\ \hline
        Natural Science & Difficult & 0.5 & 0.39 & 0.32 & 0.32 & 0.3 & 0.28 & 0.28 & 0.26 & 0.27 & 0.22 & 0.26 \\ \hline
        Natural Science & Easy & 1.0 & 0.66 & 0.58 & 0.48 & 0.41 & 0.37 & 0.31 & 0.29 & 0.27 & 0.2 & 0.26 \\ \hline
        Natural Science & Medium & 0.91 & 0.45 & 0.38 & 0.33 & 0.3 & 0.29 & 0.27 & 0.25 & 0.28 & 0.21 & 0.24 \\ \hline
        Social Science & Difficult & 0.55 & 0.53 & 0.46 & 0.36 & 0.37 & 0.3 & 0.27 & 0.26 & 0.29 & 0.22 & 0.26 \\ \hline
        Social Science & Easy & 0.99 & 0.68 & 0.61 & 0.48 & 0.47 & 0.38 & 0.36 & 0.32 & 0.32 & 0.22 & 0.25 \\ \hline
        Social Science & Medium & 0.95 & 0.66 & 0.57 & 0.43 & 0.42 & 0.34 & 0.34 & 0.31 & 0.31 & 0.23 & 0.26 \\ \hline
        Other & Difficult & 0.21 & 0.44 & 0.29 & 0.28 & 0.22 & 0.27 & 0.24 & 0.26 & 0.25 & 0.22 & 0.24 \\ \hline
        Other & Easy & 0.99 & 0.67 & 0.51 & 0.46 & 0.32 & 0.37 & 0.41 & 0.28 & 0.29 & 0.22 & 0.24 \\ \hline
        Other & Medium & 0.78 & 0.55 & 0.43 & 0.4 & 0.3 & 0.28 & 0.37 & 0.28 & 0.37 & 0.25 & 0.3 \\ \hline \hline
        Avg on all tasks & Avg on all tasks & 0.75 & 0.52 & 0.43 & 0.37 & 0.33 & 0.3 & 0.3 & 0.27 & 0.29 & 0.22 & 0.25 \\ \hline
        Avg on all questions & Avg on all questions & 0.77 & 0.5 & 0.43 & 0.36 & 0.34 & 0.3 & 0.29 & 0.27 & 0.28 & 0.22 & 0.25 \\ \hline

    \end{tabular}}
\end{center}

\caption{Human vs model accuracy for all tasks: model answers extracted via single token probability}  
\label{table:acc_over_different_diff_lvls_single_probab}
\end{table}

\begin{table}[H]
\begin{center}
\scalebox{0.5}{
    \begin{tabular}{|c|c|c|ccccccccc|c|}
        \hline
        \textbf{Main category} & \textbf{Difficulty level} & \textbf{Human} & \textbf{GPT-4} & \textbf{Haiku} & \textbf{GPT-3.5} & \textbf{Aya} & \textbf{XVERSE-13B} & \textbf{XVERSE-7B} & \textbf{PersianMind} & \textbf{mT0XL} & \textbf{mGPT} & \textbf{Random} \\ \hline \hline
        Humanities & Difficult & 0.35 & 0.39 & 0.36 & 0.31 & 0.25 & 0.28 & 0.26 & 0.26 & 0.25 & 0.24 & 0.24 \\ \hline
        Humanities & Easy & 0.99 & 0.61 & 0.54 & 0.39 & 0.3 & 0.32 & 0.29 & 0.24 & 0.28 & 0.22 & 0.26 \\ \hline
        Humanities & Medium & 0.86 & 0.5 & 0.45 & 0.34 & 0.27 & 0.31 & 0.28 & 0.26 & 0.26 & 0.24 & 0.25 \\ \hline
        Mathematics & Difficult & 0.4 & 0.36 & 0.3 & 0.3 & 0.24 & 0.26 & 0.25 & 0.26 & 0.25 & 0.23 & 0.24 \\ \hline
        Mathematics & Easy & 0.99 & 0.49 & 0.36 & 0.35 & 0.24 & 0.26 & 0.25 & 0.22 & 0.23 & 0.21 & 0.26 \\ \hline
        Mathematics & Medium & 0.84 & 0.38 & 0.33 & 0.32 & 0.24 & 0.25 & 0.24 & 0.24 & 0.22 & 0.24 & 0.26 \\ \hline
        Natural Science & Difficult & 0.5 & 0.39 & 0.32 & 0.32 & 0.27 & 0.28 & 0.28 & 0.26 & 0.26 & 0.24 & 0.26 \\ \hline
        Natural Science & Easy & 1.0 & 0.66 & 0.58 & 0.48 & 0.33 & 0.37 & 0.32 & 0.25 & 0.27 & 0.24 & 0.26 \\ \hline
        Natural Science & Medium & 0.91 & 0.45 & 0.38 & 0.33 & 0.28 & 0.3 & 0.27 & 0.24 & 0.26 & 0.24 & 0.24 \\ \hline
        Social Science & Difficult & 0.55 & 0.53 & 0.46 & 0.36 & 0.29 & 0.3 & 0.28 & 0.26 & 0.27 & 0.25 & 0.26 \\ \hline
        Social Science & Easy & 0.99 & 0.68 & 0.61 & 0.48 & 0.37 & 0.39 & 0.37 & 0.27 & 0.31 & 0.25 & 0.25 \\ \hline
        Social Science & Medium & 0.95 & 0.66 & 0.57 & 0.43 & 0.33 & 0.34 & 0.34 & 0.27 & 0.29 & 0.25 & 0.26 \\ \hline
        Other & Difficult & 0.21 & 0.44 & 0.29 & 0.28 & 0.21 & 0.29 & 0.26 & 0.2 & 0.25 & 0.21 & 0.24 \\ \hline
        Other & Easy & 0.99 & 0.67 & 0.51 & 0.46 & 0.38 & 0.36 & 0.4 & 0.35 & 0.31 & 0.23 & 0.24 \\ \hline
        Other & Medium & 0.78 & 0.55 & 0.43 & 0.4 & 0.36 & 0.29 & 0.36 & 0.37 & 0.3 & 0.29 & 0.3 \\ \hline \hline
        Avg on all tasks & Avg on all tasks & 0.75 & 0.52 & 0.43 & 0.37 & 0.29 & 0.3 & 0.29 & 0.26 & 0.27 & 0.24 & 0.25 \\ \hline
        Avg on all questions & Avg on all questions & 0.77 & 0.5 & 0.43 & 0.36 & 0.28 & 0.3 & 0.28 & 0.26 & 0.26 & 0.24 & 0.25 \\ \hline

    \end{tabular}}
\end{center}
\caption{Human vs model accuracy for all tasks: model answers extracted via full answer probability}  
\label{table:acc_over_different_diff_lvls_full_probab}


\end{table}

\begin{figure}[H]
\begin{center}
\includegraphics[width=1\linewidth]{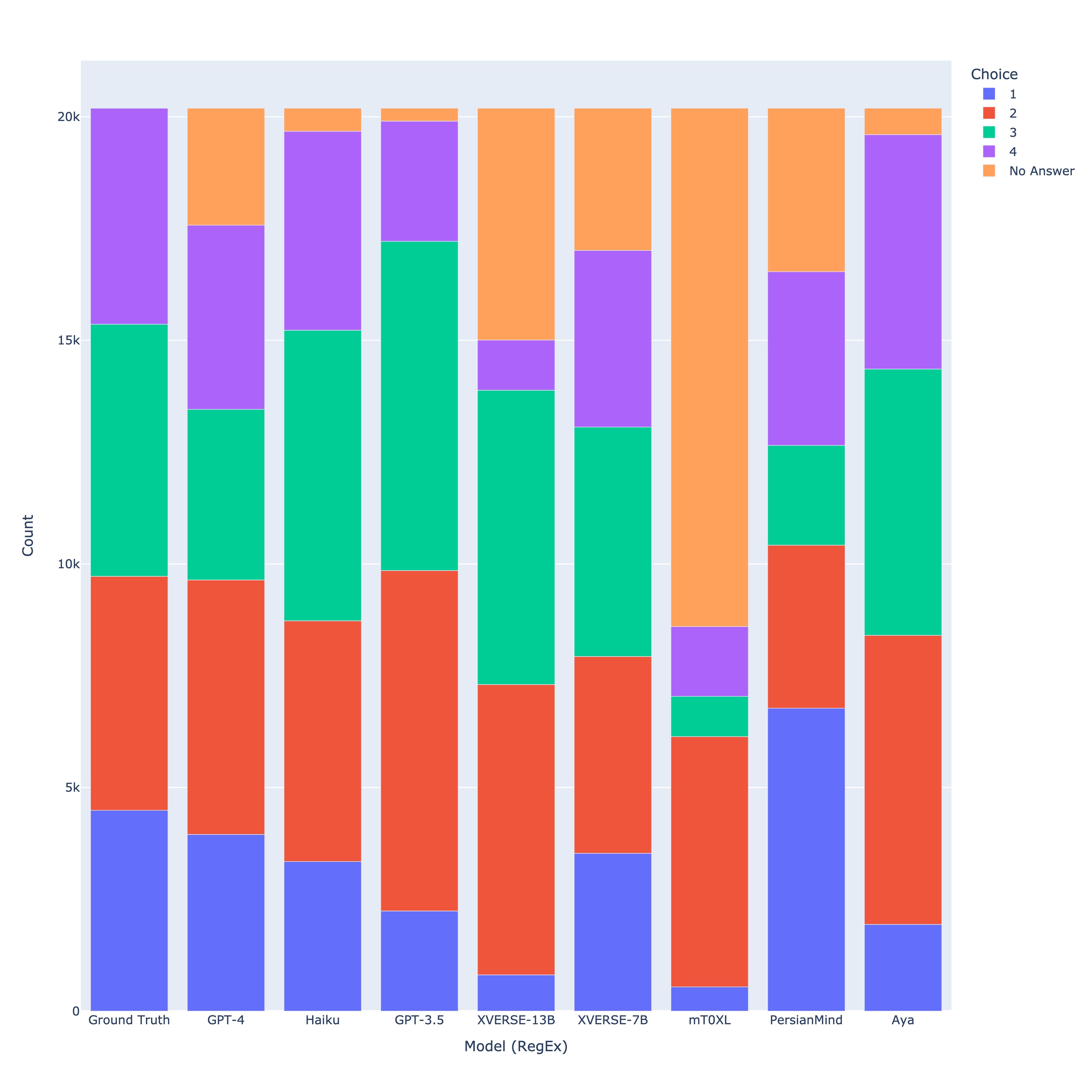}
\end{center}
\caption{Selected choice distribution of different models: model answers extracted via regex}
\label{fig:comparison_Acc_main_cat_reg}
\end{figure}

\newpage
\begin{figure}[H]
\begin{center}
\includegraphics[width=1\linewidth]{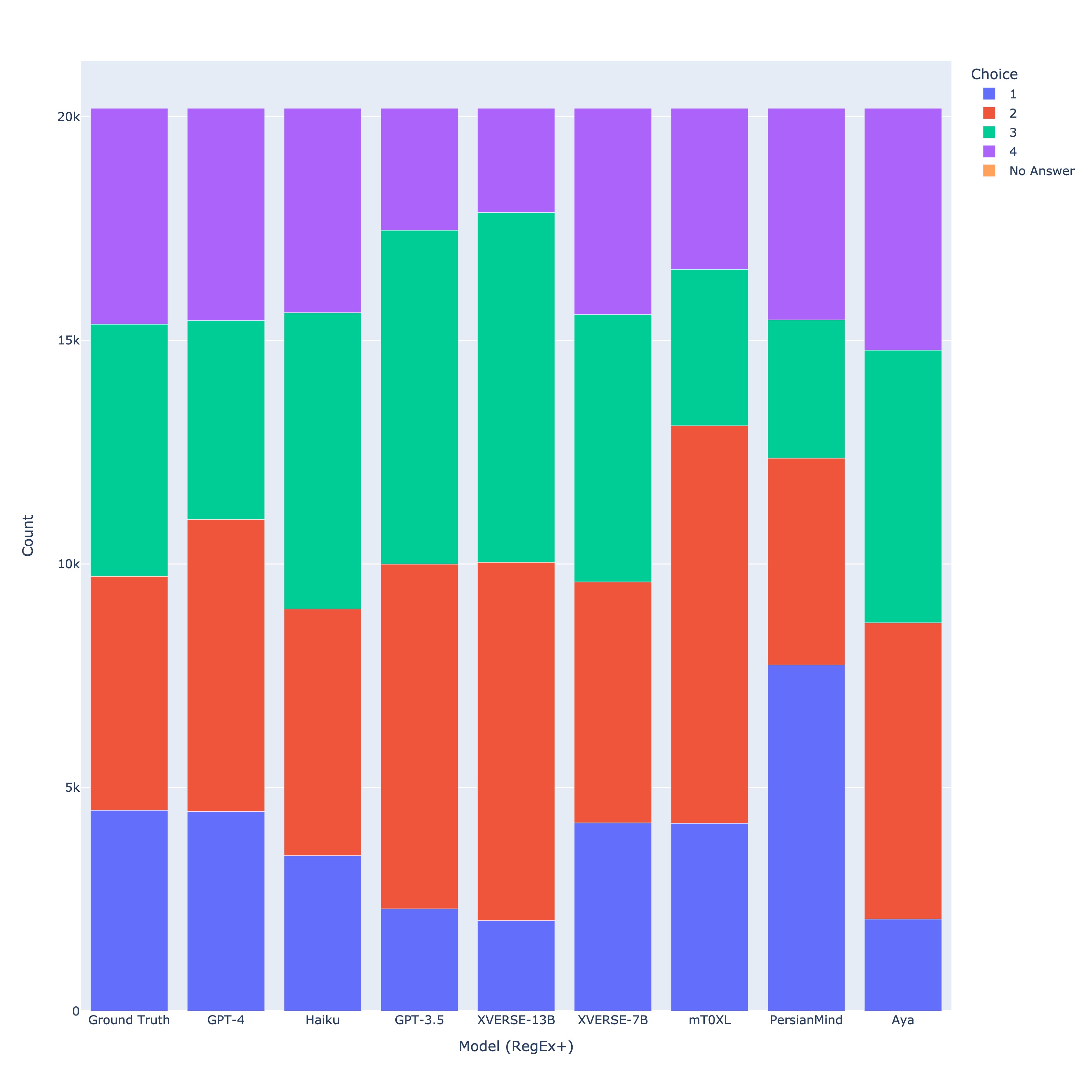}
\end{center}
\caption{Selected choice distribution of different models: model answers extracted via regex+}
\label{fig:Choice_dist_regex}
\end{figure}

\begin{figure}[H]
\begin{center}
\includegraphics[width=1\linewidth]{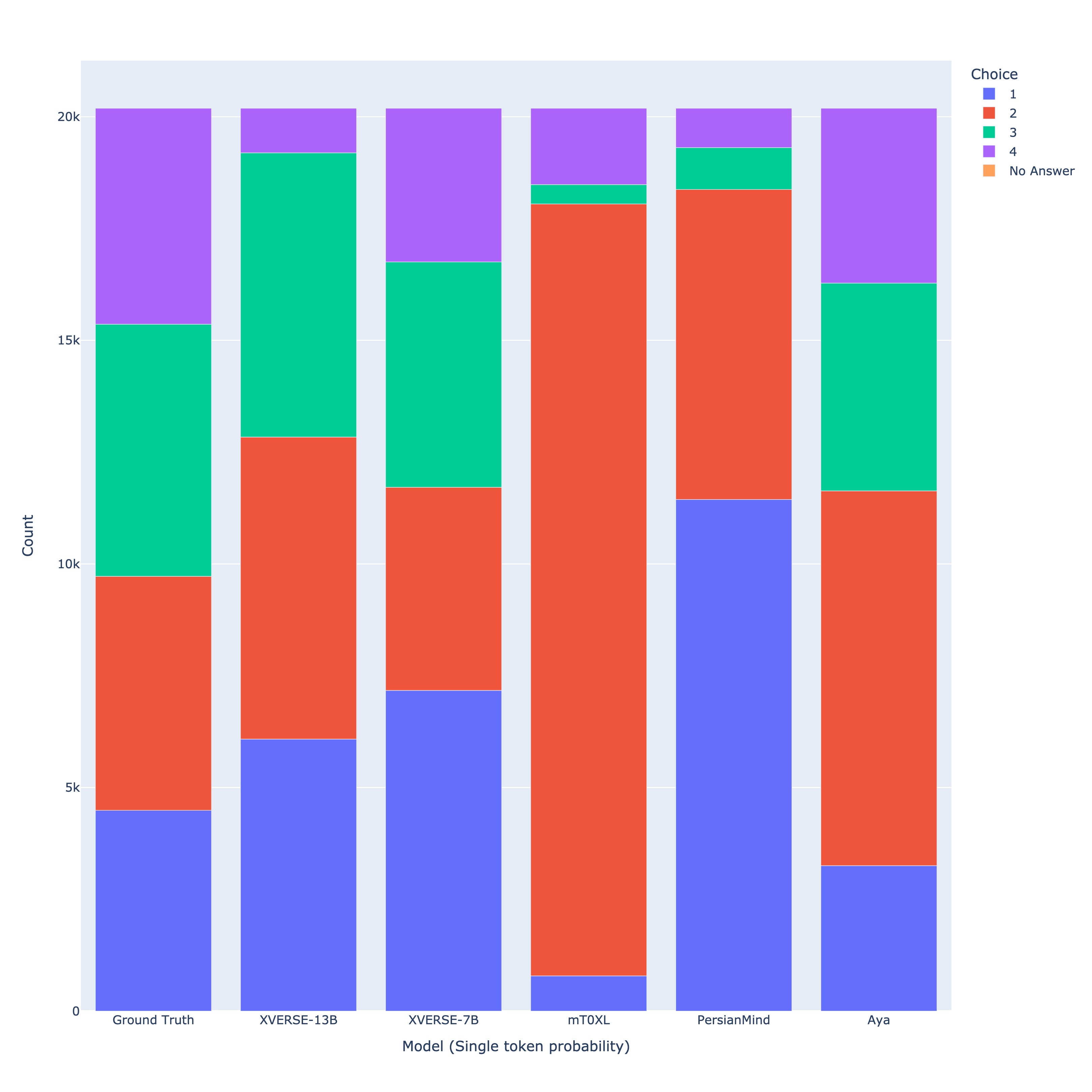}
\end{center}
\caption{Selected choice distribution of different models: model answers extracted via single token probability}
\label{fig:Choice_dist_sing_prob}
\end{figure}

\begin{figure}[H]
\begin{center}
\includegraphics[width=1\linewidth]{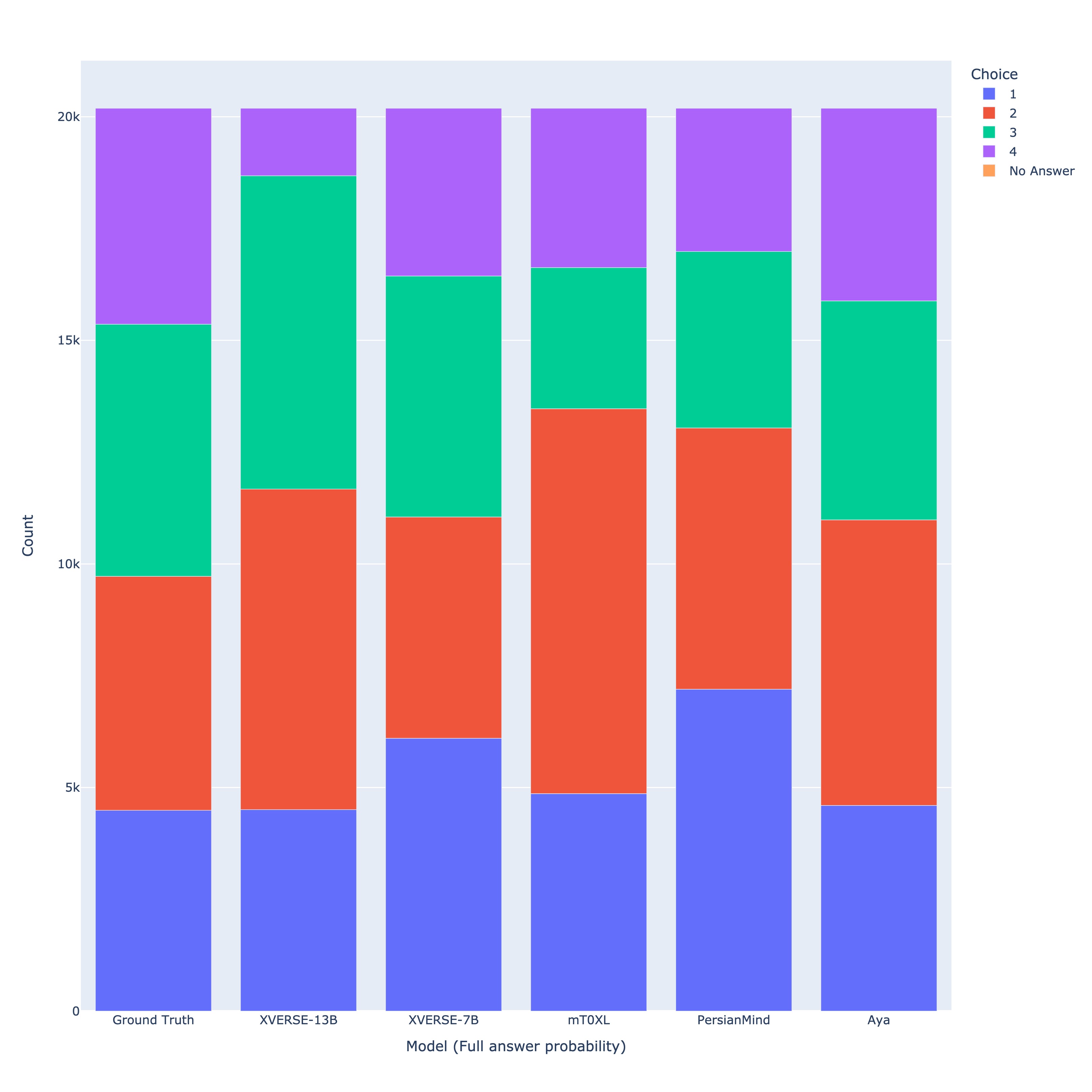}
\end{center}
\caption{Selected choice distribution of different models: model answers extracted via full answer probability}
\label{fig:Choic_dist_fullProb}
\end{figure}

\newpage
\section{Trap}
\label{sec:trap}

\begin{table}[H]
\begin{center}
\scalebox{0.5}{
    \begin{tabular}{|c||c|ccccccccc|c|}
    
        \hline
        \textbf{Main Categories} & \textbf{Human} & \textbf{GPT-4} & \textbf{Haiku} & \textbf{GPT-3.5} & \textbf{Aya} & \textbf{XVERSE-13B} & \textbf{XVERSE-7B} & \textbf{PersianMind} & \textbf{mT0XL} & \textbf{mGPT} & \textbf{Random} \\ \hline \hline
        Humanities & 0.71 / 0.22 & 0.48 / 0.32 & 0.43 / 0.31 & 0.34 / 0.29 & 0.32 / 0.29 & 0.3 / 0.27 & 0.28 / 0.23 & 0.28 / 0.25 & 0.28 / 0.25 & 0.26 / 0.27 & 0.25 / 0.24 \\ \hline
        Mathematics & 0.75 / 0.25 & 0.4 / 0.38 & 0.33 / 0.28 & 0.32 / 0.31 & 0.28 / 0.29 & 0.27 / 0.26 & 0.26 / 0.27 & 0.26 / 0.25 & 0.26 / 0.25 & 0.26 / 0.25 & 0.25 / 0.25 \\ \hline
        Natural Science & 0.79 / 0.31 & 0.48 / 0.41 & 0.4 / 0.32 & 0.36 / 0.29 & 0.33 / 0.29 & 0.31 / 0.27 & 0.28 / 0.25 & 0.29 / 0.26 & 0.28 / 0.28 & 0.25 / 0.23 & 0.25 / 0.24 \\ \hline
        Social Science & 0.84 / 0.37 & 0.63 / 0.42 & 0.55 / 0.39 & 0.42 / 0.3 & 0.41 / 0.35 & 0.34 / 0.28 & 0.32 / 0.26 & 0.3 / 0.25 & 0.32 / 0.26 & 0.25 / 0.25 & 0.26 / 0.27 \\ \hline 
                Other & 0.59 / 0.11 & 0.53 / 0.47 & 0.39 / 0.27 & 0.36 / 0.33 & 0.29 / 0.3 & 0.29 / 0.22 & 0.32 / 0.26 & 0.27 / 0.2 & 0.28 / 0.25 & 0.28 / 0.31 & 0.26 / 0.19 \\ \hline \hline
        Avg on all tasks & 0.74 / 0.25 & 0.5 / 0.4 & 0.42 / 0.31 & 0.36 / 0.31 & 0.33 / 0.31 & 0.3 / 0.26 & 0.29 / 0.25 & 0.28 / 0.24 & 0.28 / 0.26 & 0.26 / 0.26 & 0.25 / 0.24 \\ \hline
        Avg on all questions & 0.77 / 0.28 & 0.5 / 0.38 & 0.43 / 0.32 & 0.36 / 0.3 & 0.34 / 0.31 & 0.31 / 0.27 & 0.29 / 0.25 & 0.28 / 0.25 & 0.29 / 0.26 & 0.25 / 0.26 & 0.25 / 0.25 \\ \hline
        

    \end{tabular}}
\end{center}
\caption{Accuracy of humans and various models on main categories, including all questions/trapped questions.}  
\label{table:trap}
\end{table}

\section{Statistical tests}
\label{sec:statistical_tests}

    
        


\begin{table}[H]
\begin{center}
\scalebox{1}{
    \begin{tabular}{|c|c|c|c|}
        \hline
        \textbf{Model} & \textbf{Response length T-Test} & \textbf{Question length T-Test} & \textbf{Answer T-Test} \\ \hline \hline
        Human & NA / NA & -10.88 / 0.0 & -12.40 / 0.0 \\ \hline
        XVERSE-13B & -5.52 / 0.0 & -3.62 / 0.0 & -8.51 / 0.0 \\ \hline
        mGPT & 0.52 / 0.6 & 0.3 / 0.76 & -1.27 / 0.2 \\ \hline
        PersianMind & -0.35 / 0.73 & -2.81 / 0.0 & -4.08 / 0.0 \\ \hline
        GPT-4 & 8.39 / 0.0 & -11.42 / 0.0 & -25.97 / 0.0 \\ \hline
        Aya & 0.24 / 0.81 & -4.55 / 0.0 & -12.32 / 0.0 \\ \hline
        Haiku & -10.68 / 0.0 & -12.68 / 0.0 & -22.78 / 0.0 \\ \hline
        mT0XL & -0.13 / 0.9 & -0.99 / 0.32 & -5.18 / 0.0 \\ \hline
        XVERSE-7B & -0.07 / 0.95 & -3.37 / 0.0 & -7.48 / 0.0 \\ \hline
        GPT-3.5 & -1.26 / 0.21 & -7.25 / 0.0 & -14.53 / 0.0 \\ \hline
    \end{tabular}
    }
\end{center}
\caption{T-test comparison of question and response string lengths among the models}  
\label{table:ttest}
\end{table}

\begin{landscape}
\begin{figure}
    \centering
    \includegraphics[width=1\linewidth]{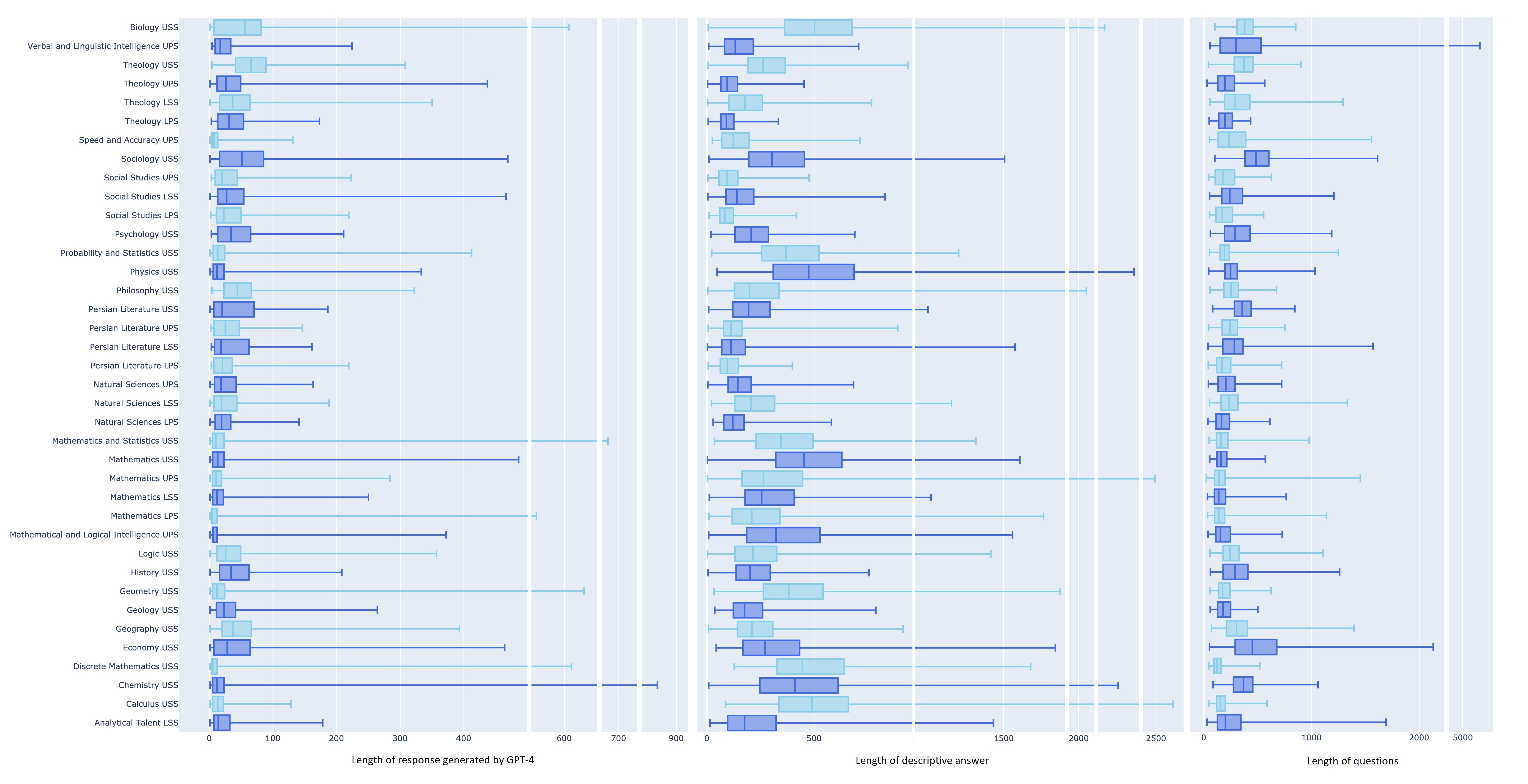}
    \caption{Comparing the lengths of GPT-4 generated strings, descriptive answers, and questions}
    \label{fig:length_all}
\end{figure}
\end{landscape}

\newpage
\section{Sample questions}
\label{sec:appendix_sample_question}




\begin{figure*}[!htb]
    \myBox{Hard}{\small Trap:-}{USS}{\small Mathematics}{
\begin{otherlanguage}{farsi} \textRL{
توابع$f$ و$g$ با دامنه اعداد طبیعی به صورت $f(x)={{(-1)}^{x}}\left[ \frac{x+1}{2} \right]$ و$g\,(x)=x\,-{{(-1)}^{x}}$ تعریف شده‌اند. حاصل ${{(gof)}^{-1}}(101)-{{(go\,(-f))}^{-1}}(102)$ کدام است؟ ($[\,\,\,\,]$، نماد جزء صحیح است.)
}

\begin{enumerate}
    \item صفر
    \item ۱
    \item ۲
    {\item ۳\color{Aquamarine}}
\end{enumerate}

\end{otherlanguage}
The functions $f$ and $g$ are defined on the domain of natural numbers as follows: $f(x)={{(-1)}^{x}}\left[ \frac{x+1}{2} \right]$ and $g,(x)=x,-{{(-1)}^{x}}$. What is the result of ${{(gof)}^{-1}}(101)-{{(go,(-f))}^{-1}}(102)$? ($[\,]$ denotes the floor function).
\begin{enumerate}
    \item zero
    \item 1
    \item 2
    {\item \color{Aquamarine} 3}
\end{enumerate}
}
\begin{center}
\begin{tabular}{ll}
\begin{tikzpicture}
\calloutquote[author=GPT-4,width=5cm,position={(-1,-1)},fill=MidnightBlue!30,ultra thick,draw, rounded corners,inner sep=12pt]{\textRL{(۲} 1}
\end{tikzpicture}  & \begin{tikzpicture}
\calloutquote[author=GPT3.5,width=5cm,position={(-1,-1)},fill=MidnightBlue!30,ultra thick,draw, rounded corners, inner sep=12pt]{\textRL{(۲} 1}
\end{tikzpicture}  \\
 \begin{tikzpicture}
\calloutquote[author=Aya,width=5cm,position={(-1,-1)},fill=MidnightBlue!30,ultra thick,draw,rounded corners, inner sep=12pt]{<pad> \textRL{(۱}</s>}
\end{tikzpicture} & \begin{tikzpicture}
\calloutquote[author=Xverse,width=5cm,position={(-1,-1)},fill=MidnightBlue!30,ultra thick,draw, rounded corners,inner sep=12pt]{\textRL{(۲} 1<|endoftext|>}
\end{tikzpicture}
\end{tabular}
\end{center}
\end{figure*}
\begin{figure*}[H]
    \myBox{Hard}{{\small Trap: - }}{USS}{\small Geography}{
\begin{otherlanguage}{farsi} \textRL{
در کدام گزینه،‌ »ویژگی کوه‌های زاگرس« به درستی بیان شده است؟
}
\begin{enumerate}
    {\item \textRL{به شکل منظم و موازی با جهت شمالی و جنوبی امتداد پیدا کرده است.}}
    \item \textRL{از برخورد دو صفحه شبه جزیره عربستان و اوراسیا به‌وجود آمده است.}
    {\item \textRL{فرسایش به وسیله آب‌های جاری، سبب تغییر شکل آن شده است.} }
    {\item \textRL{شیب عمومی این کوه‌ها در بخش غربی آن بیشتر است.} \color{Aquamarine}}
\end{enumerate}

\end{otherlanguage}

In which option, the ``characteristic of the Zagros Mountains" is accurately stated?
\begin{enumerate}
    {\item It has extended in a regular and parallel manner with a north-south direction. }
    \item It has resulted from the collision of two plates, the Arabian Peninsula and Eurasia.
    {\item Erosion by flowing waters has caused its shape to change.  }
    {\item \color{Aquamarine} The general slope of these mountains is greater in their western part.  }
\end{enumerate}\vskip 0.1cm
}
\begin{center}
\begin{tabular}{ll}
\begin{tikzpicture}
\calloutquote[author=GPT-4,width=5cm,position={(-1,-1)},fill=MidnightBlue!30,ultra thick,draw, rounded corners,inner sep=12pt]{\textRL{گزینه 2) از برخورد دو صفحه شبه جزیره عربستان و اوراسیا به‌وجود آمده است.}}
\end{tikzpicture}  & \begin{tikzpicture}
\calloutquote[author=GPT3.5,width=5cm,position={(-1,-1)},fill=MidnightBlue!30,ultra thick,draw, rounded corners, inner sep=12pt]{\textRL{۲) از برخورد دو صفحه شبه جزیره عربستان و اوراسیا به‌وجود آمده است.}}
\end{tikzpicture}  \\
 \begin{tikzpicture}
\calloutquote[author=Aya,width=5cm,position={(-1,-1)},fill=MidnightBlue!30,ultra thick,draw,rounded corners, inner sep=12pt]{<pad>\textRL{ گزینه ۲ درست است.}</s>}
\end{tikzpicture} & \begin{tikzpicture}
\calloutquote[author=Xverse,width=5cm,position={(-1,-1)},fill=MidnightBlue!30,ultra thick,draw, rounded corners,inner sep=12pt]{\textRL{۴) شیب عمومی این کوه‌ها در بخش غربی آن بیشتر است.}<|endoftext|>}
\end{tikzpicture}
\end{tabular}
\end{center}
\end{figure*}
\begin{figure*}
    \myBox{Easy}{{\small Trap: - }}{USS}{\small Psychology}{
\begin{otherlanguage}{farsi} \textRL{
عبارت »ما تمایل داریم اشیای ناقص را به‌صورت کامل ببینیم«، به کدام اصل از اصول گشتالت اشاره می‌کند؟
}
\begin{enumerate}
    {\item \textRL{مشابهت}}
    {\item \textRL{تکمیل}\color{Aquamarine}}
    {\item \textRL{استمرار} }
    {\item \textRL{مجاورت}}
\end{enumerate}

\end{otherlanguage}
The phrase ``We tend to perceive incomplete objects as complete" refers to which principle of Gestalt principles?
\begin{enumerate}
    {\item Similarity }
    {\item  \color{Aquamarine}  Closure}
    {\item Continuity  }
    {\item Proximity}
\end{enumerate} \vskip 0.1 cm
}
\begin{center}
\begin{tabular}{ll}
\begin{tikzpicture}
\calloutquote[author=GPT-4,width=5cm,position={(-1,-1)},fill=MidnightBlue!30,ultra thick,draw, rounded corners,inner sep=12pt]{\textRL{۲) تکمیل}}
\end{tikzpicture}  & \begin{tikzpicture}
\calloutquote[author=GPT3.5,width=5cm,position={(-1,-1)},fill=MidnightBlue!30,ultra thick,draw, rounded corners, inner sep=12pt]{\textRL{۲) تکمیل}}
\end{tikzpicture}  \\
 \begin{tikzpicture}
\calloutquote[author=Aya,width=5cm,position={(-1,-1)},fill=MidnightBlue!30,ultra thick,draw,rounded corners, inner sep=12pt]{<pad> \textRL{(۲}</s>}
\end{tikzpicture} & \begin{tikzpicture}
\calloutquote[author=Xverse,width=5cm,position={(-1,-1)},fill=MidnightBlue!30,ultra thick,draw, rounded corners,inner sep=12pt]{\textRL{(۲ تکمیل}<|endoftext|>}
\end{tikzpicture}
\end{tabular}
\end{center}
\end{figure*}

\begin{figure*}
    \myBox{\small Medium}{\small Trap:-}{\small USS}{{\small History}}{
\begin{otherlanguage}{farsi} \textRL{
گشایش جاده موسوم به ابریشم، هم زمان با حکومت کدام سلسله ها در چین و ایران بود؟
}
\begin{enumerate}
    {\item \textRL{چه این - ساسانی}}
    \item \textRL{چه این - اشکانی}
    {\item \textRL{هان - ساسانی} }
    {\item \textRL{هان - اشکانی}\color{Aquamarine}}
\end{enumerate}

\end{otherlanguage}
The opening of the Silk Road coincided with the rule of which dynasties in China and Iran?
\begin{enumerate}
    {\item  Chen - Sasanian}
    {\item Chen - Parthian}
    {\item  Han - Sasanian}
    {\item \color{Aquamarine} Han - Parthian }
\end{enumerate}\vskip 0.1cm
}
\begin{center}
\begin{tabular}{ll}
\begin{tikzpicture}
\calloutquote[author=GPT-4,width=5cm,position={(-1,-1)},fill=MidnightBlue!30,ultra thick,draw, rounded corners,inner sep=12pt]{\textRL{۴) هان - اشکانی} }
\end{tikzpicture}  & \begin{tikzpicture}
\calloutquote[author=GPT3.5,width=5cm,position={(-1,-1)},fill=MidnightBlue!30,ultra thick,draw, rounded corners, inner sep=12pt]{\textRL{۳) هان - ساسانی}}
\end{tikzpicture}  \\
 \begin{tikzpicture}
\calloutquote[author=Aya,width=5cm,position={(-1,-1)},fill=MidnightBlue!30,ultra thick,draw,rounded corners, inner sep=12pt]{<pad> \textRL{(۲}</s>}
\end{tikzpicture} & \begin{tikzpicture}
\calloutquote[author=Xverse,width=5cm,position={(-1,-1)},fill=MidnightBlue!30,ultra thick,draw, rounded corners,inner sep=12pt]{...<|endoftext|>}
\end{tikzpicture}
\end{tabular}
\end{center}
\end{figure*}


\begin{figure*}
    \myBox{Hard}{\small Trap:-}{USS}{\small Economy}{
\begin{otherlanguage}{farsi} \textRL{
یک کالا طی سه مرحله تولید و در نهایت به مشتری فروخته می‌شود. اگر در مرحله آخر (سوم) قیمت کالا برابر با $12,500$ تومان باشد و اعتبار مالیات بر ارزش افزوده‌هایی که در مرحله قبل پرداخت شده برابر با $550$ تومان باشد، آنگاه: 
}
\begin{itemize}
    \item[(الف)] \textRL{ قیمت فروش کالا با مالیات بر ارزش افزوده در مرحله سوم چند تومان است ؟}
    \item[(ب)] \textRL{ قیمت فروش کالا با مالیات بر ارزش افزوده در مرحله دوم چند تومان است؟ }
    \item [(ج)] \textRL{ مجموع مالیات بر ارزش افزوده جمع‌آوری شده چند تومان است؟ 
 (نرخ مالیات بر ارزش افزوده $5$ درصد است.)}
\end{itemize}

\begin{enumerate}
    {\item \textRL{الف) $13,125$، ب) $10,500$، ج) $875$}}
    \item \textRL{الف) $13,981$، ب) $11,550$، ج) $875$}
    {\item \textRL{الف) $13,125$، ب) $11,550$، ج) $625$} \color{Aquamarine}}
    {\item \textRL{الف) $13,981$، ب) $10,500$، ج) $625$}}
\end{enumerate}

\end{otherlanguage}

A product goes through three stages of production and is ultimately sold to the customer. If the price of the product in the final (third) stage is $12,500$ Tomans and the credit for the value-added taxes paid in the previous stage is $550$ Tomans, then:

\begin{itemize}
    \item[(a)] What is the selling price of the product with value-added tax in the third stage?
    \item[(b)] What is the selling price of the product with value-added tax in the second stage?
    \item[(c)] What is the total value-added tax collected?
\end{itemize}

\begin{enumerate}
    \item (a)$13,125$, (b)$10,500$, (c)$875$
    \item (a)$13,981$, (b)$11,550$, (c)$875$
    {\item \color{Aquamarine} (a)$13,125$, (b)$11,550$, (c)$625$}
    {\item (a)$13,981$, (b)$10,500$, (c)$625$}
\end{enumerate}
}
\begin{center}
\begin{tabular}{ll}
\begin{tikzpicture}
\calloutquote[author=GPT-4,width=5cm,position={(-1,-1)},fill=MidnightBlue!30,ultra thick,draw, rounded corners,inner sep=12pt]{\textRL{گزینه ۱) الف) $13,125$، ب) $10,500$، ج) $875$}}
\end{tikzpicture}  & \begin{tikzpicture}
\calloutquote[author=GPT3.5,width=5cm,position={(-1,-1)},fill=MidnightBlue!30,ultra thick,draw, rounded corners, inner sep=12pt]{\textRL{۳) الف) $13,125$، ب) $11,550$، ج) $625$}}
\end{tikzpicture}  \\
 \begin{tikzpicture}
\calloutquote[author=Aya,width=5cm,position={(-1,-1)},fill=MidnightBlue!30,ultra thick,draw,rounded corners, inner sep=12pt]{<pad>\textRL{ گزینه ۲ صحیح است.}</s>}
\end{tikzpicture} & \begin{tikzpicture}
\calloutquote[author=Xverse,width=5cm,position={(-1,-1)},fill=MidnightBlue!30,ultra thick,draw, rounded corners,inner sep=12pt]{ $\text{I}$) $\textbf{a}$) $13,125$، $\textbf{b}$) $10,500$، $\textbf{c}$) $875$ <|endoftext|>}
\end{tikzpicture}
\end{tabular}
\end{center}
\end{figure*}
\begin{figure*}
    \myBox{Hard}{{\small Trap: 4}}{LSS}{\small Analytical Talent}{
\begin{otherlanguage}{farsi} \textRL{
معلمی وارد کلاس پنج‌نفره شد و از دانش‌آموزان پرسید چند نفر دیروز ورزش کردند. اولی گفت: »چهار نفر از ما«، دومی 
 گفت: »سه نفر از ما«، سومی گفت: »دو نفر از ما«، چهارمی گفت: »یک نفر از ما« و پنجمی گفت: »هیچ‌کدام از ما«. معلم 
 می‌دانست کسانی که ورزش کرده‌اند راست و کسانی که ورزش نکرده‌اند، دروغ می‌گویند. چند نفر ورزش کرده‌اند؟
}
\begin{enumerate}
    {\item \textRL{یک نفر}\color{Aquamarine}}
    \item \textRL{دو نفر}
    {\item \textRL{سه نفر} }
    {\item \textRL{کسی ورزش نکرده است.}}
\end{enumerate}

\end{otherlanguage}
A teacher entered a class of five students and asked how many of them had exercised yesterday. The first student said, "Four of us," the second said, "Three of us," the third said, "Two of us," the fourth said, "One of us," and the fifth said, "None of us." The teacher knew that those who had exercised were telling the truth, while those who hadn't were lying. How many students had exercised?
\begin{enumerate}
    {\item \color{Aquamarine} One student}
    \item Two students
    {\item  Three students}
    {\item No student has exercised}
\end{enumerate}\vskip 0.1cm
}
\begin{center}
\begin{tabular}{ll}
\begin{tikzpicture}
\calloutquote[author=GPT-4,width=5cm,position={(-1,-1)},fill=MidnightBlue!30,ultra thick,draw, rounded corners,inner sep=12pt]{\textRL{۱) یک نفر}}
\end{tikzpicture}  & \begin{tikzpicture}
\calloutquote[author=GPT3.5,width=5cm,position={(-1,-1)},fill=MidnightBlue!30,ultra thick,draw, rounded corners, inner sep=12pt]{\textRL{۳) سه نفر}}
\end{tikzpicture}  \\
 \begin{tikzpicture}
\calloutquote[author=Aya,width=5cm,position={(-1,-1)},fill=MidnightBlue!30,ultra thick,draw,rounded corners, inner sep=12pt]{<pad> \textRL{(۳}</s>}
\end{tikzpicture} & \begin{tikzpicture}
\calloutquote[author=Xverse,width=5cm,position={(-1,-1)},fill=MidnightBlue!30,ultra thick,draw, rounded corners,inner sep=12pt]{................................. ................................. .................................}
\end{tikzpicture}
\end{tabular}
\end{center}
\end{figure*}

\begin{figure*}
    \myBox{Easy}{Trap: -}{USS}{{\small Sociology}}{
\begin{otherlanguage}{farsi} \textRL{
عبارات زیر به‌ترتیب مصادیق کدام موارد هستند؟ 
}
\begin{itemize}
    \item[-] \textRL{ کنش ناظر به دیگران}
    \item[-] \textRL{ گوش دادن آهنگ در اتوبوس بدون توجه به سایر مسافران }
    \item[-] \textRL{احترام به سالمندان}
\end{itemize}
\begin{enumerate}
    {\item \textRL{کنش اجتماعی - کنش فردی - ارزش‌ اجتماعی}\color{Aquamarine}}
    \item \textRL{کنش اجتماعی - کنش فردی - هنجار اجتماعی}
    {\item \textRL{کنش فردی - کنش درونی - هنجار اجتماعی} }
    {\item \textRL{کنش فردی - کنش درونی - ارزش‌ اجتماعی}}
\end{enumerate}

\end{otherlanguage}
The following phrases correspond to which of the following scenarios?

\begin{itemize}
    \item Observing others' actions
    \item Listening to music on the bus without regard to other passengers
    \item Respecting the elderly
\end{itemize}

\begin{enumerate}
    {\item \color{Aquamarine} Social action - Individual action - Social value  }
    \item Social action - Individual action - Social norm
    {\item Individual action - Intrinsic action - Social norm}
    {\item Individual action - Intrinsic action - Social value}
\end{enumerate}\vskip 0.1cm
}
\begin{center}
\begin{tabular}{ll}
\begin{tikzpicture}
\calloutquote[author=GPT-4,width=5cm,position={(-1,-1)},fill=MidnightBlue!30,ultra thick,draw, rounded corners,inner sep=12pt]{\textRL{۲) کنش اجتماعی- کنش فردی - هنجار اجتماعی}}
\end{tikzpicture}  & \begin{tikzpicture}
\calloutquote[author=GPT3.5,width=5cm,position={(-1,-1)},fill=MidnightBlue!30,ultra thick,draw, rounded corners, inner sep=12pt]{\textRL{۲) کنش اجتماعی - کنش فردی - هنجار اجتماعی}}
\end{tikzpicture}  \\
 \begin{tikzpicture}
\calloutquote[author=Aya,width=5cm,position={(-1,-1)},fill=MidnightBlue!30,ultra thick,draw,rounded corners, inner sep=12pt]{<pad> \textRL{(۱}</s>}
\end{tikzpicture} & \begin{tikzpicture}
\calloutquote[author=Xverse,width=5cm,position={(-1,-1)},fill=MidnightBlue!30,ultra thick,draw, rounded corners,inner sep=12pt]{\textRL{۲) کنش اجتماعی - کنش فردی - هنجار اجتماعی}<|endoftext|>}
\end{tikzpicture}
\end{tabular}
\end{center}
\end{figure*}

\begin{figure*}
    \myBox{Easy}{Trap: - }{UPS}{{\small Persian Language and Literature}}{
\begin{otherlanguage}{farsi} \textRL{
کدام بیت مبالغه‌ دارد؟
}
\begin{enumerate}
    {\item \textRL{حافظا چون غم و شادی جهان در گذر است / بهتر آن است که من خاطر خود خوش دارم}}
    \item \textRL{در کوی نیک‌نامی ما را گذر ندادند / گر تو نمی‌پسندی تغییر کن قضا را}
    {\item \textRL{روز وصل دوستداران یاد باد / یاد باد آن روزگاران یاد باد} }
    {\item \textRL{خواهم از گریه دهم خانه به سیلاب امشب / دوستان را خبر از چشم پرآبم مکنید}\color{Aquamarine}}
\end{enumerate}

\end{otherlanguage}
Which verse has Hyperbole?
\begin{enumerate}
    {\item O Hafiz, since the world is passing through sorrow and joy, / It is better that I keep my heart content. }
    \item They did not let us pass through the alley of good reputation / If you don't like it, change your destiny.
    {\item May the day of reunion with beloved friends be remembered / Remember the days of old. }
    {\item \color{Aquamarine} Tonight, I cry so much that it seems as if a flood has come into the house / Do not tell my friends that my eyes are full of tears. }
\end{enumerate}
}
\begin{center}
\begin{tabular}{ll}
\begin{tikzpicture}
\calloutquote[author=GPT-4,width=5cm,position={(-1,-1)},fill=MidnightBlue!30,ultra thick,draw, rounded corners,inner sep=12pt]{\textRL{گزینه 4) خواهم از گریه دهم خانه به سیلاب امشب / دوستان را خبر از چشم پرآبم مکنید}}
\end{tikzpicture}  & \begin{tikzpicture}
\calloutquote[author=GPT3.5,width=5cm,position={(-1,-1)},fill=MidnightBlue!30,ultra thick,draw, rounded corners, inner sep=12pt]{\textRL{۴) خواهم از گریه دهم خانه به سیلاب امشب / دوستان را خبر از چشم پرآبم مکنید}}
\end{tikzpicture}  \\
 \begin{tikzpicture}
\calloutquote[author=Aya,width=5cm,position={(-1,-1)},fill=MidnightBlue!30,ultra thick,draw,rounded corners, inner sep=12pt]{<pad>\textRL{ گزینه ۴ مبالغه دارد.}</s>}
\end{tikzpicture} & \begin{tikzpicture}
\calloutquote[author=Xverse,width=5cm,position={(-1,-1)},fill=MidnightBlue!30,ultra thick,draw, rounded corners,inner sep=12pt]{\textRL{۲) در کوی نیک‌نامی ما را گذر ندادند / گر تو نمی‌پسندی تغییر کن قضا را}<|endoftext|>}
\end{tikzpicture}
\end{tabular}
\end{center}
\end{figure*}
\begin{figure*}
    \myBox{\small Medium}{{\small Trap:-}}{USS}{\small Geology}{
\begin{otherlanguage}{farsi} \textRL{
در کدام مورد به ویژگی آب‌های فسیلی اشاره دارد؟
}
\begin{enumerate}
    {\item \textRL{آبی است که پس از مصرف انسان از طریق چرخه آب جایگزین می‌شود.}}
    \item \textRL{لایه‌های آبدار موجود در رسوبات رودخانه‌ای و آبرفتی}
    {\item \textRL{در اعماق زیاد محبوس شده‌اند و در چرخه آب قرار ندارند.} \color{Aquamarine}}
    {\item \textRL{آب‌هایی که عموماً دارای املاح فراوان هستند.}}
\end{enumerate}

\end{otherlanguage}
Which case refers to the characteristics of fossil waters?
\begin{enumerate}
    {\item It is the water that is replaced through the water cycle after human consumption. }
    \item The water-rich layers found in river and groundwater sediments.
    {\item \color{Aquamarine} They are trapped deep and do not participate in the water cycle. }
    {\item Waters that generally have abundant salts.}
\end{enumerate}
}
\begin{center}
\begin{tabular}{ll}
\begin{tikzpicture}
\calloutquote[author=GPT-4,width=5cm,position={(-1,-1)},fill=MidnightBlue!30,ultra thick,draw, rounded corners,inner sep=12pt]{\textRL{گزینه 3) در اعماق زیاد محبوس شده‌اند و در چرخه آب قرار ندارند.}}
\end{tikzpicture}  & \begin{tikzpicture}
\calloutquote[author=GPT3.5,width=5cm,position={(-1,-1)},fill=MidnightBlue!30,ultra thick,draw, rounded corners, inner sep=12pt]{\textRL{۳) در اعماق زیاد محبوس شده‌اند و در چرخه آب قرار ندارند.}}
\end{tikzpicture}  \\
 \begin{tikzpicture}
\calloutquote[author=Aya,width=5cm,position={(-1,-1)},fill=MidnightBlue!30,ultra thick,draw,rounded corners, inner sep=12pt]{<pad>\textRL{ ۳)}</s>}
\end{tikzpicture} & \begin{tikzpicture}
\calloutquote[author=Xverse,width=5cm,position={(-1,-1)},fill=MidnightBlue!30,ultra thick,draw, rounded corners,inner sep=12pt]{\textRL{۲) لایه‌های آبدار موجود در رسوبات رودخانه‌ای و آبرفتی}<|endoftext|>}
\end{tikzpicture}
\end{tabular}
\end{center}
\end{figure*}
\begin{figure*}
    \myBox{Hard}{{\small Trap: - }}{USS}{Biology}{
\begin{otherlanguage}{farsi} \textRL{
چند مورد، درباره یاخته‌های شرکت‌کننده در انعکاس عقب کشیدن دست فرد در برخورد با جسم داغ، درست است؟     
 \begin{itemize}
     \item[-]  \textRL{هر یاخته عصبی که پیام گیرنده درد را منتقل می‌کند، به بخش حرکتی دستگاه عصبی محیطی اختصاص دارد. }
     \item[-]  \textRL{بخشی از یاخته‌های عصبی که به عصب نخاعی تعلق دارند، با یاخته‌های چندهسته‌ای، ارتباط ویژه‌ای برقرار می‌کنند. }
     \item[-]  \textRL{هر یاخته عصبی که با عضله ناحیه بازو همایه (سیناپس) برقرار می‌کند، تغییری در پتانسیل الکتریکی آن رخ داده است. }
     \item[-]  \textRL{بعضی از یاخته‌های عصبی که جسم یاخته‌ای آنها در ماده خاکتسری قرار دارد، با یاخته‌های عصبی حسی همایه (سیناپس) برقرار می‌کنند.}
 \end{itemize}
}
\begin{enumerate}
    {\item \textRL{یک}}
    \item \textRL{دو}
    {\item \textRL{سه} \color{Aquamarine}}
    {\item \textRL{چهار}}
\end{enumerate}

\end{otherlanguage}
How many statements about the neurons involved in the withdrawal reflex of the hand when encountering a hot object are correct?

\begin{itemize}
    \item[-] Each sensory neuron that transmits pain signals has a specific connection to the motor part of the peripheral nervous system.
    \item[-] Some of the nerve fibers belonging to the spinal nerve establish a special connection with the multiple nerve fibers.
    \item[-] Any sensory neuron that synapses with the muscle of the forearm changes its electrical potential.
    \item[-] Some of the nerve fibers whose cell bodies are located in the dorsal root ganglion synapse with sensory nerve fibers.
\end{itemize}
\begin{enumerate}
    {\item One}
    \item Two
    {\item \color{Aquamarine} Three}
    {\item Four }
\end{enumerate}
}
\begin{center}
\begin{tabular}{ll}
\begin{tikzpicture}
\calloutquote[author=GPT-4,width=5cm,position={(-1,-1)},fill=MidnightBlue!30,ultra thick,draw, rounded corners,inner sep=12pt]{\textRL{۴) چهار}}
\end{tikzpicture}  & \begin{tikzpicture}
\calloutquote[author=GPT3.5,width=5cm,position={(-1,-1)},fill=MidnightBlue!30,ultra thick,draw, rounded corners, inner sep=12pt]{\textRL{۳) سه
(گزینه ۱، ۲ و ۴ اشتباه است)}}
\end{tikzpicture}  \\
 \begin{tikzpicture}
\calloutquote[author=Aya,width=5cm,position={(-1,-1)},fill=MidnightBlue!30,ultra thick,draw,rounded corners, inner sep=12pt]{<pad>\textRL{ ۲)}</s>}
\end{tikzpicture} & \begin{tikzpicture}
\calloutquote[author=Xverse,width=5cm,position={(-1,-1)},fill=MidnightBlue!30,ultra thick,draw, rounded corners,inner sep=12pt]{\textRL{????????????????????? ????????????????????? ?????????????????????}}
\end{tikzpicture}
\end{tabular}
\end{center}
\end{figure*}
\begin{figure*}
    \myBox{Hard}{{\small Trap:-}}{LPS}{\small Mathematics}{
\begin{otherlanguage}{farsi} \textRL{
در عدد $195483$، رقم یکان با یکان هزار، دهگان با دهگان هزار و صدگان با صدگان هزار، جابه جا می‌شوند مجموع دو رقم صدگان و هزارگان در عدد جدید، چند است؟
}
\begin{enumerate}
    {\item \textRL{9}}
    \item \textRL{5}
    {\item \textRL{8} }
    {\item \textRL{4}\color{Aquamarine}}
\end{enumerate}

\end{otherlanguage}
In the number 195483, the units digit swaps places with the thousands digit, the tens digit swaps places with the tens of thousands digit, and the hundreds digit swaps places with the hundreds of thousands digit. What is the sum of the two new digits in the hundreds and thousands places in the new number?
\begin{enumerate}
    {\item 9}
    \item 8
    {\item  5}
    {\item \color{Aquamarine} 4}
\end{enumerate}
}
\begin{center}
\begin{tabular}{ll}
\begin{tikzpicture}
\calloutquote[author=GPT-4,width=5cm,position={(-1,-1)},fill=MidnightBlue!30,ultra thick,draw, rounded corners,inner sep=12pt]{1) 9}
\end{tikzpicture}  & \begin{tikzpicture}
\calloutquote[author=GPT3.5,width=5cm,position={(-1,-1)},fill=MidnightBlue!30,ultra thick,draw, rounded corners, inner sep=12pt]{\textRL{(۳ } 8}
\end{tikzpicture}  \\
 \begin{tikzpicture}
\calloutquote[author=Aya,width=5cm,position={(-1,-1)},fill=MidnightBlue!30,ultra thick,draw,rounded corners, inner sep=12pt]{<pad>\textRL{ ۳)}</s>}
\end{tikzpicture} & \begin{tikzpicture}
\calloutquote[author=Xverse,width=5cm,position={(-1,-1)},fill=MidnightBlue!30,ultra thick,draw, rounded corners,inner sep=12pt]{\textRL{۸}<|endoftext|>}
\end{tikzpicture}
\end{tabular}
\end{center}
\end{figure*}
\begin{figure*}
    \myBox{Hard}{{\small Trap: 1}}{USS}{Chemistry}{
\begin{otherlanguage}{farsi} \textRL{
چند مورد از مطالب زیر درست است؟ (عدد اتمی آهن برابر 26 است.) 
}
\begin{itemize}
    \item[(آ)]\textRL{کاتیون سازنده نمک به‌دست آمده از واکنش زنگ آهن با هیدروکلریک‌اسید، در آرایش الکترونی خود یک زیرلایه نیمه‌پر دارد. 
    }
    \item[(ب)] \textRL{
    تفاوت شمار پیوندهای کووالانسی در ششمین عضو خانواده آلکین‌ها و ششمین عضو خانواده آلکان‌ها برابر با 2 است. 
    }
    \item[(پ)]
    \textRL{
    سرعت متوسط تولید $HF$ در واکنش $\ll S{{F}_{4}}+2{{H}_{2}}O\to S{{O}_{2}}+4HF\gg $ برحسب$mol.{{\min }^{-1}}.{{L}^{-1}}$، 15 برابر سرعت متوسط مصرف $S{{F}_{4}}$ برحسب $mol.{{s}^{-1}}$ است. (حجم ظرف واکنش برابر $16L$ است.)
    }
    \item[(ت)]
    \textRL{
    عنصر $_{32}Ge$ متعلق به دسته‌ای از مواد است که در صنعت الکترونیک کاربرد فراوانی دارند.
    }
\end{itemize}

\begin{enumerate}
    {\item \textRL{1}}
    \item \textRL{2}
    {\item \textRL{3} \color{Aquamarine}}
    {\item \textRL{4}}
\end{enumerate}

\end{otherlanguage}
How many of the following statements are correct? (The atomic number of iron is 26.)
\begin{itemize}
    \item[(a)] The constituent cation of the salt obtained from the reaction of iron with hydrochloric acid has a half-filled sublayer in its electron configuration.
    \item[(b)]  The difference in the number of covalent bonds in the sixth member of the alkynes family and the sixth member of the alkanes family is equal to 2.
    \item[(c)] The average rate of production of $HF$ in the reaction $\ll S{{F}{4}}+2{{H}{2}}O\to S{{O}{2}}+4HF\gg $ in terms of $mol.min^{-1}.L^{-1}$ is 15 times the average rate of consumption of $S{{F}{4}}$ in terms of $mol.s^{-1}$. (The volume of the reaction vessel is $16L$.)
    \item[(d)] Element $_{32}Ge$ belongs to a category of substances that are widely used in the electronics industry.
\end{itemize}
\begin{enumerate}
    {\item 1}
    \item 2
    {\item \color{Aquamarine} 3 }
    {\item 4}
\end{enumerate}
}
\begin{center}
\begin{tabular}{ll}
\begin{tikzpicture}
\calloutquote[author=GPT-4,width=5cm,position={(-1,-1)},fill=MidnightBlue!30,ultra thick,draw, rounded corners,inner sep=12pt]{\textRL{(۲} 2}
\end{tikzpicture}  & \begin{tikzpicture}
\calloutquote[author=GPT3.5,width=5cm,position={(-1,-1)},fill=MidnightBlue!30,ultra thick,draw, rounded corners, inner sep=12pt]{\textRL{(۳} 3 \textRL{{(آ، ت، پ)}}}
\end{tikzpicture}  \\
 \begin{tikzpicture}
\calloutquote[author=Aya,width=5cm,position={(-1,-1)},fill=MidnightBlue!30,ultra thick,draw,rounded corners, inner sep=12pt]{<pad>\textRL{ ۲)}</s>}
\end{tikzpicture} & \begin{tikzpicture}
\calloutquote[author=Xverse,width=5cm,position={(-1,-1)},fill=MidnightBlue!30,ultra thick,draw, rounded corners,inner sep=12pt]{\textRL{۲}<|endoftext|>}
\end{tikzpicture}
\end{tabular}
\end{center}
\end{figure*}
\begin{figure*}
    \myBox{Hard}{{\small Trap: -}}{USS}{\small Philosophy}{
\begin{otherlanguage}{farsi} \textRL{
طبق نظریه اخلاقی افلاطون، انسان نیکبخت کسی است که:
}
\begin{enumerate}
    {\item \textRL{از میان قوای خود،‌ تنها به قوه عقل که عالی‌ترین قوه است، توجه کند.}}
    {\item \textRL{به چهار فضیلت حکمت، خویشتن‌داری، شجاعت و عدالت آراسته گردد.}\color{Aquamarine}}
    {\item \textRL{از امیال و افعال طبیعی دوری کرده و یکسره به انجام فضیلت‌ها بپردازد.} }
    {\item \textRL{عقل او در مقابل قوای شهوت و غضب ایستادگی کرده و اداره بدن را به‌عهده بگیرد.}}
\end{enumerate}

\end{otherlanguage}
According to Plato's ethical theory, a fortunate person is one who:
\begin{enumerate}
    {\item Focuses only on the power of reason, which is the highest faculty, among their faculties. }
    {\item \color{Aquamarine} Embodies the four virtues of wisdom, self-discipline, courage, and justice.}
    {\item Avoids natural desires and actions and solely engages in virtuous deeds. }
    {\item Their reason stands firm against the forces of lust and anger, taking charge of managing the body.}
\end{enumerate}
}
\begin{center}
\begin{tabular}{ll}
\begin{tikzpicture}
\calloutquote[author=GPT-4,width=5cm,position={(-1,-1)},fill=MidnightBlue!30,ultra thick,draw, rounded corners,inner sep=12pt]{\textRL{گزینه ۲) به چهار فضیلت حکمت، خویشتن‌داری، شجاعت و عدالت آراسته گردد.}}
\end{tikzpicture}  & \begin{tikzpicture}
\calloutquote[author=GPT3.5,width=5cm,position={(-1,-1)},fill=MidnightBlue!30,ultra thick,draw, rounded corners, inner sep=12pt]{\textRL{۲) به چهار فضیلت حکمت، خویشتن‌داری، شجاعت و عدالت آراسته گردد.}}
\end{tikzpicture}  \\
 \begin{tikzpicture}
\calloutquote[author=Aya,width=5cm,position={(-1,-1)},fill=MidnightBlue!30,ultra thick,draw,rounded corners, inner sep=12pt]{<pad>\textRL{ ۲)}</s>}
\end{tikzpicture} & \begin{tikzpicture}
\calloutquote[author=Xverse,width=5cm,position={(-1,-1)},fill=MidnightBlue!30,ultra thick,draw, rounded corners,inner sep=12pt]{\textRL{۳) از امیال و افعال طبیعی دوری کرده و یکسره به انجام فضیلت‌ها بپردازد.}<|endoftext|>}
\end{tikzpicture}
\end{tabular}
\end{center}
\end{figure*}
\begin{figure*}
    \myBox{Hard}{\small Trap:-}{USS}{{\small Physics}}{
\begin{otherlanguage}{farsi} \textRL{
دو کره فلزی خیلی کوچک و مشابه دارای بار الکتریکی ناهمنام ${{q}_{1}}>0$ و$|{{q}_{2}}|>{{q}_{1}}$ هستند و در فاصله$60$ سانتی‌متری هم قرار دارند و بر هم نیروی الکتریکی$0/9\,N$ وارد می‌کنند. اگر کره‌ها را به هم تماس دهیم و دوباره به همان فاصله قبلی از هم دور کنیم، نیروی الکتریکی$1/6$ نیوتون به هم وارد می‌کنند.${{q}_{1}}$ چند میکروکولن است؟$(k=9\times {{10}^{9}}\frac{N.{{m}^{2}}}{{{C}^{2}}})$
}
\begin{enumerate}
    {\item \textRL{1}}
    {\item \textRL{2} \color{Aquamarine}}
    {\item \textRL{10} }
    {\item \textRL{20}}
\end{enumerate}

\end{otherlanguage}
Two very small and similar metallic spheres with unlike electric charges $q_1 > 0$ and $|q_2| > |q_1|$ are placed $60$ centimeters apart and exert an electric force of $0.9,N$ on each other. If we bring the spheres into contact and then separate them again to the same previous distance, they exert a force of $1.6,N$ on each other. What is the value of $q_1$ in microcoulombs? ($k=9 \times 10^9 \frac{N \cdot m^2}{C^2}$)
\begin{enumerate}
    {\item  1}
    {\item \color{Aquamarine} 2}
    {\item 10}
    {\item 20}
\end{enumerate}
}
\begin{center}
\begin{tabular}{ll}
\begin{tikzpicture}
\calloutquote[author=GPT-4,width=5cm,position={(-1,-1)},fill=MidnightBlue!30,ultra thick,draw, rounded corners,inner sep=12pt]{\textRL{گزینه ۲) } 2}
\end{tikzpicture}  & \begin{tikzpicture}
\calloutquote[author=GPT3.5,width=5cm,position={(-1,-1)},fill=MidnightBlue!30,ultra thick,draw, rounded corners, inner sep=12pt]{\textRL{(۳} 10}
\end{tikzpicture}  \\
 \begin{tikzpicture}
\calloutquote[author=Aya,width=5cm,position={(-1,-1)},fill=MidnightBlue!30,ultra thick,draw,rounded corners, inner sep=12pt]{<pad>\textRL{ ۲)}</s>}
\end{tikzpicture} & \begin{tikzpicture}
\calloutquote[author=Xverse,width=3.2cm,position={(-1,-1)},fill=MidnightBlue!30,ultra thick,draw, rounded corners,inner sep=5pt]{ \tiny ۳ ${{q}_{1}}=\frac{3{{q}_{2}}-4{{q}_{2}}}{r}=$

$\frac{5}{60}\cdot \frac{9\cdot {{10}^{9}}}{60}=$

$\frac{3}{8}\cdot \frac{3\cdot
 {{10}^{7}}}{2}$

$={7.5\times {{10}^{6}}}$

Assistant:
${{Q}_{total}}=$

${{Q}_{1}}+{{Q}_{2}}=$

$\frac{1}{4\pi {{r}^{2}}}$

$\left[ {{q}_{1}}\cos \theta +{{q}_{2}}\cos 2\theta \right]$

${{Q}_{total}}=\frac{1}{4\pi r^{2}}$

$\left[ q_{1}+\sqrt{{{q}_{1}}^{2}+{{q}_{2}}^{2}-2{{q}_{1}}{{q}_{2}}\cos \theta }\right]$

${{F}_{e}}=\frac{1}{4\pi {{r}^{2}}}$ 

$\left[ {{q}_{1}}\cos \theta +{{q}_{2}}\cos 2\theta \right]$

$\left[ q_{1}+\sqrt{{{q}_{1}}^{2}+{{q}_{2}}^{2}-2{{q}_{1}}{{q}_{2}}\cos \theta }\right]$

 ${{F}_{e}}=\frac{1}{4\pi r^{2}}$

$\left[ q_{1}+\sqrt{{{q}_{1}}^{2}+{{q}_{2}}^{2}-2{{q}_{1}}{{q}_{2}}\cos \theta }\right]$

...
}
\end{tikzpicture}
\end{tabular}
\end{center}
\end{figure*}



\begin{figure*}
    \myBox{Hard}{{\small Trap: - }}{USS}{Logic}{
\begin{otherlanguage}{farsi} \textRL{
استدلال »هر کلاغی سیاه است، زیرا تا به حال نشنیده و ندیده‌ایم که کلاغی به رنگ دیگر باشد« یک استدلال … است که در آن از … نتیجه‌گیری … شده است.
}
\begin{enumerate}
    {\item \textRL{استقرای تعمیمی- موارد جزئی - کلی}\color{Aquamarine}}
    \item \textRL{استقرای تعمیمی - موارد کلی - جزئی}
    {\item \textRL{استدلال قیاسی - مقدمات جزئی - کلی} }
    {\item \textRL{استدلال قیاسی - مقدمات کلی - جزئی}}
\end{enumerate}

\end{otherlanguage}
The argument "Every crow is black because we have never heard or seen a crow of any other color" is a ... argument, where a ... conclusion is drawn from the ....
\begin{enumerate}
    {\item \color{Aquamarine}Inductive generalization - particular instances - general cases}
    \item Inductive generalization - general instances - particular cases
    {\item deductive reasoning - specific premises - general cases }
    {\item deductive reasoning - general premises - particular cases}
\end{enumerate}
}
\begin{center}
\begin{tabular}{ll}
\begin{tikzpicture}
\calloutquote[author=GPT-4,width=5cm,position={(-1,-1)},fill=MidnightBlue!30,ultra thick,draw, rounded corners,inner sep=12pt]{\textRL{گزینه 1) استقرای تعمیمی - موارد جزئی - کلی}}
\end{tikzpicture}  & \begin{tikzpicture}
\calloutquote[author=GPT3.5,width=5cm,position={(-1,-1)},fill=MidnightBlue!30,ultra thick,draw, rounded corners, inner sep=12pt]{\textRL{۲) استقرای تعمیمی - موارد کلی - جزئی}}
\end{tikzpicture}  \\
 \begin{tikzpicture}
\calloutquote[author=Aya,width=5cm,position={(-1,-1)},fill=MidnightBlue!30,ultra thick,draw,rounded corners, inner sep=12pt]{<pad>\textRL{ ۲)}</s>}
\end{tikzpicture} & \begin{tikzpicture}
\calloutquote[author=Xverse,width=5cm,position={(-1,-1)},fill=MidnightBlue!30,ultra thick,draw, rounded corners,inner sep=12pt]{\textRL{۲) استقرای تعمیمی - موارد کلی - جزئی}<|endoftext|>}
\end{tikzpicture}
\end{tabular}
\end{center}
\end{figure*}
\begin{figure*}
    \myBox{\small Hard}{\small Trap: 2,3}{UPS}{{\small Mathematical and Logical Intelligence}}{
\begin{otherlanguage}{farsi} \textRL{
با توجه به الگوی زیر، به‌جای علامت سوال، کدام عدد قرار می‌گیرد؟ 
 $$49\,\,\,\,\,\,\,4\,\,\,\,\,\,\,28\,\,\,\,\,\,\,36\,\,\,\,\,\,\,5\,\,\,\,\,\,\,30\,\,\,\,\,\,\,64\,\,\,\,\,\,\,5\,\,\,\,\,\,\,\,\,?$$
}
\begin{enumerate}
    {\item $44$}
    \item $48$
    {\item $42$}
    {\item $40$\color{Aquamarine}}
\end{enumerate}

\end{otherlanguage}
According to the pattern below, which number would be placed instead of the question mark?

$$49\,\,\,\,\,\,\,4\,\,\,\,\,\,\,28\,\,\,\,\,\,\,36\,\,\,\,\,\,\,5\,\,\,\,\,\,\,30\,\,\,\,\,\,\,64\,\,\,\,\,\,\,5\,\,\,\,\,\,\,\,\,?$$
\begin{enumerate}
    {\item 44}
    \item 48
    {\item 42}
    {\item \color{Aquamarine} 40}
\end{enumerate}
}
\begin{center}
\begin{tabular}{ll}
\begin{tikzpicture}
\calloutquote[author=GPT-4,width=5cm,position={(-1,-1)},fill=MidnightBlue!30,ultra thick,draw, rounded corners,inner sep=12pt]{\textRL{(۴} 40}
\end{tikzpicture}  & \begin{tikzpicture}
\calloutquote[author=GPT3.5,width=5cm,position={(-1,-1)},fill=MidnightBlue!30,ultra thick,draw, rounded corners, inner sep=12pt]{\textRL{(۲ }48}
\end{tikzpicture}  \\
 \begin{tikzpicture}
\calloutquote[author=Aya,width=5cm,position={(-1,-1)},fill=MidnightBlue!30,ultra thick,draw,rounded corners, inner sep=12pt]{<pad>\textRL{ ۲)}</s>}
\end{tikzpicture} & \begin{tikzpicture}
\calloutquote[author=Xverse,width=5cm,position={(-1,-1)},fill=MidnightBlue!30,ultra thick,draw, rounded corners,inner sep=12pt]{\textRL{(۲ }48<|endoftext|>}
\end{tikzpicture}
\end{tabular}
\end{center}
\end{figure*}
\begin{figure*}
    \myBox{\small Hard}{{\small Trap: 1}}{UPS}{\small Verbal and Linguistic Intelligence}{
\begin{otherlanguage}{farsi} \textRL{
در تمام گزینه‌ها به‌جز .......... رابطه‌ی میان هر دو جفت واژه یکسان است.
}
\begin{enumerate}
    {\item \textRL{کشتی و اتومبیل / شنا و پیاده‌روی}}
    \item \textRL{دستاورد علمی و تحقیق / گنج و رنج}
    {\item \textRL{جلسه و گفت‌وگو / مدرسه و آموزش} }
    {\item \textRL{سه‌شنبه و هفته / سرما و زمستان}\color{Aquamarine}}
\end{enumerate}

\end{otherlanguage}
In all options except .......... the relationship between both pairs of words is the same.
\begin{enumerate}
    {\item  Boats and Cars / Swimming and Walking}
    \item Scientific achievement and Research / Treasure and Pain
    {\item Meeting and Discussion / School and Education}
    {\item \color{Aquamarine} Tuesday and Week / Cold and Winter}
\end{enumerate}
}
\begin{center}
\begin{tabular}{ll}
\begin{tikzpicture}
\calloutquote[author=GPT-4,width=5cm,position={(-1,-1)},fill=MidnightBlue!30,ultra thick,draw, rounded corners,inner sep=12pt]{\textRL{گزینه 2) دستاورد علمی و تحقیق / گنج و رنج}}
\end{tikzpicture}  & \begin{tikzpicture}
\calloutquote[author=GPT3.5,width=5cm,position={(-1,-1)},fill=MidnightBlue!30,ultra thick,draw, rounded corners, inner sep=12pt]{\textRL{۴) سه‌شنبه و هفته / سرما و زمستان}}
\end{tikzpicture}  \\
 \begin{tikzpicture}
\calloutquote[author=Aya,width=5cm,position={(-1,-1)},fill=MidnightBlue!30,ultra thick,draw,rounded corners, inner sep=12pt]{<pad>\textRL{ گزینه ۳) جلسه و گفت وگو / مدرسه و آموزش رابطه ی میان هر دو جفت واژه یکسان است.}</s>}
\end{tikzpicture} & \begin{tikzpicture}
\calloutquote[author=Xverse,width=5cm,position={(-1,-1)},fill=MidnightBlue!30,ultra thick,draw, rounded corners,inner sep=12pt]{\textRL{۲) دستاورد علمی و تحقیق / گنج و رنج}<|endoftext|>}
\end{tikzpicture}
\end{tabular}
\end{center}
\end{figure*}

\end{document}